\documentclass[journal]{IEEEtran}

\usepackage{amsmath,bm}
\usepackage{cite}
\usepackage{nicealgo}
\usepackage[numbers,sort&compress]{natbib}
\usepackage{etoolbox}
\usepackage{multicol,xcolor,color,colortbl,hhline,caption}
\usepackage[linesnumbered,ruled]{algorithm2e}

\usepackage{graphicx,url}

\begin{document}

\title{A Novel Approach for Optimum-Path Forest Classification Using Fuzzy Logic}

\author{Renato~W.~R.~de~Souza, Jo\~{a}o V. C. de Oliveira,% 
	~Leandro A. Passos,~\IEEEmembership{Member,~IEEE}, Weiping Ding,~\IEEEmembership{Senior Member,~IEEE}, Jo\~{a}o P. Papa,~\IEEEmembership{Senior Member,~IEEE} and Victor Hugo C. de Albuquerque,~\IEEEmembership{Senior Member,~IEEE}
        
\thanks{Renato William R. de Souza and Victor Hugo C. de Albuquerque are with the Graduate Program in Applied Informatics, University of Fortaleza, Fortaleza/CE, Brazil.(email: victor.albuquerque@unifor.br and renatowilliam@edu.unifor.br)
}% <-this % stops a space
\thanks{Jo\~ao Vitor Chaves de Oliveira is with the Pontifical Catholic University of Rio de Janeiro, Brazil.
(email: joliveira@inf.puc-rio.br).}% <-this % stops a space
\thanks{Leandro A. Passos and Jo\~{a}o P. Papa are with S\~ao Paulo State University, Brazil. 
(email: leandropassosjr@gmail.com and joao.papa@unesp.br).}% <-this % stops a space
\thanks{Weiping Ding is with the School of Information Science and Technology, Nantong University, Nantong, China. (email:ding.wp@ntu.edu.cn) (\emph{Corresponding author.})}% <-this % stops a space

\thanks{Manuscript received xxxx.}}

\markboth{IEEE TRANSACTIONS ON FUZZY SYSTEMS}%
{Souza \MakeLowercase{\textit{et al.}}: Fuzzy Optimum-Path Forest}

\maketitle

\begin{abstract}
In the past decades, fuzzy logic has played an essential role in many research areas. Alongside, graph-based  pattern recognition has shown to be of great importance due to its flexibility in partitioning the feature space using the background from graph theory. Some years ago, a new framework for both supervised, semi-supervised, and unsupervised learning named Optimum-Path Forest (OPF) was proposed with competitive results in several applications, besides comprising a low computational burden. In this paper, we propose the Fuzzy Optimum-Path Forest, an improved version of the standard OPF classifier that learns the samples' membership in an unsupervised fashion, which are further incorporated during supervised training. Such information is used to identify the most relevant training samples, thus improving the classification step. Experiments conducted over twelve public datasets highlight the robustness of the proposed approach, which behaves similarly to standard OPF in worst-case scenarios.
\end{abstract}


\IEEEpeerreviewmaketitle

\section{Introduction}
\label{s.introduction}

\IEEEPARstart{M}achine learning techniques achieved notorious popularity in the last decades due to the outstanding success while solving complex problems in a large variety of fields, such as image and character recognition, medical diagnosis, and remote sensing, among others. Such techniques are usually classified into \emph{supervised learning}, i.e., algorithms whose training step employs a labeled set of samples to find the parameters that properly fit the model, and \emph{unsupervised learning}, which clusters the data accordingly to their similarity among each other.

Among these techniques, the so-called Optimum-Path Forest (OPF) has achieved remarkable results in the last years in a broad range of applications. OPF is a graph-based framework proposed to tackle supervised~\cite{PapaIJIST:09,PapaPR:2012,PapaPRL:17}, unsupervised~\cite{RochaIJIST:09} and also semi-supervised problems~\cite{AmorimPR:16,AmorimIS:18}. The reason for such success lies on five main aspects: (i) the OPF training step is usually much faster than traditional machine learning approaches, (ii) it does not require a proper selection of hyperparameters for some variants, (iii) it deals with a large volume of data efficiently; (iv) it does not assume separability of samples in the feature space, and (v) its unsupervised version can find clusters on-the-fly, i.e., there is no need to know the number of clusters beforehand.

Roughly speaking, OPF-based classifiers work on a reward-competition process, where some predefined key samples, i.e., \emph{prototypes}, compete among themselves to partition the graph into optimum-path trees (OPTs). Such structures may represent clusters (unsupervised learning) or a group of samples from the same class. The competition process is guided by a \emph{path-cost function} that is minimized/maximized for each dataset sample. Therefore, depending on the adjacency relation, the methodology to estimate prototypes, and path-cost function, one can design a different classifier.

In the past years, several studies aimed at improving the performance of OPF-based classifiers. Passos et al.~\cite{PASSOSEPSR:2016}, for instance, proposed an adaptation of the unsupervised OPF to deal with anomaly detection (OPF-AD) in the context of non-technical losses in electricity distribution systems. Later on, Guimar\~{a}es et al.~\cite{8515679} proposed to fine-tune OPF-AD using meta-heuristic optimization techniques in the context of anomaly detection in wireless sensor networks. 

Fernandes et al.~\cite{FernandesIEEETSG:18} proposed a modified version of the supervised OPF classifier that computes the probability of a given sample belonging to a determined class, instead of the hard classification output given by standard OPF. Fernandes et al.~\cite{FernandesPAA:17} also proposed to compute a \emph{confidence value} for each training sample to deal with plateaus during learning. Additionally, OPF was also employed for improving the performance of other algorithms, such as Neural Networks with Radial Basis Function~\cite{rosaICPR:2014} and the Brain Storm Optimization Algorithm~\cite{afonsoSACI:2018}, to cite a few.

However, as far as we are concerned, we have not observed any attempt to incorporate fuzzy logic into the OPF formalism. Besides, standard OPF does not consider the importance of samples during the training step, which makes the classifier more prone to overfitting. As a matter of fact, H\"{u}llermeier~\cite{hullermeier2005fuzzy} stated that fuzzy methods have the potential to contribute to machine learning in several ways, turning data classification faster and more robust concerning sensitivity to data variation. Based on such an assumption, many researchers employed fuzzy operators to boost a variety of models. Lin and Wang~\cite{lin2002fuzzy}, for instance, proposed the Fuzzy Support Vector Machines. Later on, Tian et al~\cite{tian2017new}, Le et al.~\cite{le2010new}, and Sevakula and Verma~\cite{sevakula2017compounding} introduced fuzzy-based SVM versions capable of reducing the computational burden and increasing accuracy rates. Several other works addressed the task of improving Restricted Boltzmann Machines using fuzzy concepts~\cite{8319509,8478774,8653841,7782438}. Moreover, fuzzy methods were recently applied in the context of deep neural networks~\cite{deng2017hierarchical}, c-means~\cite{lei2018significantly}, and transfer learning~\cite{zuo2019fuzzy}, among others. 

The main contribution of this paper is to introduce fuzzy concepts into the OPF framework by proposing a new variant, which has shown to outperform standard OPF in several datasets. Experiments conducted over twelve public and private datasets confirm the robustness of the model, which is compared against three popular techniques: Linear Support Vector Machines SVM (SVM)~\cite{platt1999probabilistic}, Naive Bayes (NB)~\cite{duda1973pattern}, and the standard OPF.

In a nutshell, the primary contributions of this paper are:

\begin{enumerate}
\item to introduce fuzzy logic in the Optimum-Path Forest formulation. Such an approach considers each sample's neighborhood to compute its membership, which is later employed in the classification process;
\item to improve naive OPF classifier by employing fuzzy concepts; and
\item to foster the literature related to fuzzy- and graph-based pattern classification.
\end{enumerate}

The remainder of this paper is organized as follows. Section~\ref{s.theoretical} describes both the supervised and unsupervised versions of the OPF, while Section~\ref{s.fopf} introduces the proposed approach. Sections~\ref{s.methodology} and~\ref{s.experiments} present the methodology and the experimental results, respectively. Finally, Section~\ref{s.conclusions} states conclusions and future works.

\section{Theoretical Background}
\label{s.theoretical}

This section presents the supervised and the unsupervised versions of the Optimum-Path Forest.

\subsection{Supervised Optimum-Path Forest}
\label{ss.sOPF}

The supervised Optimum-path Forest~\cite{PapaIJIST:09,PapaPR:2012,PapaPRL:17} is a graph-based approach that poses the task of data classification as a graph partition problem, where each node is represented by a sample (i.e., feature vector) and the edges connect every pair of nodes. The prototypes samples compete among themselves to ``conquer''\ non-prototype nodes by offering them optimum-path costs. The output of the algorithm, a collection of optimum-path trees, defines an optimum-path forest.

The training step aims at minimizing the cost assigned to each sample at the very beginning of the algorithm. The notion of \emph{optimality} is encoded by the path-cost function, which must follow some constraints, i.e., OPF requires such function to be a \emph{smooth} one~\cite{FalcaoPAMI:04}. The supervised OPF variant considered in this work~\cite{PapaIJIST:09,PapaPR:2012} employs $f_{max}$ as the path-cost function, which is computed as follows:

\begin{eqnarray}
f_{max}(\langle
\bm{q}\rangle) & = & \left\{ \begin{array}{ll}
  0 & \mbox{if $\bm{q}\in {\cal P}$,} \\
  +\infty & \mbox{otherwise}
  \end{array}\right. \nonumber \\
  f_{max}(\phi \cdot \langle \bm{q},\bm{u} \rangle) & = & \max\{f_{max}(\phi),d(\bm{q},\bm{u})\}, 
  \label{e.pathCost}
\end{eqnarray}
where $d(\bm{q},\bm{u})$ stands for the distance (i.e., arc-weight) between samples $\bm{q}$ and $\bm{u}$, ${\cal P}$ stands for the set prototypes, and $\phi$ denotes a path, i.e., a sequence of adjacent samples with no repetition\footnote{The notation $\phi \cdot \langle \bm{q},\bm{u} \rangle$ denotes the concatenation between the path $\phi$ and the edge $\langle \bm{q},\bm{u} \rangle$.}. Additionally, the maximum distance among adjacent samples in the path $\phi$ is computed by $f_{max}(\phi)$ when $\phi$ is not a trivial path. 

Papa et al.~\cite{PapaIJIST:09} proposed to find ${\cal P}^\star$ by exploring a theoretical property that guarantees zero error during training when using $f_{max}$ as the path-cost function and the adjacent samples with different classes in the training set~\cite{AlleneIVC:10}. Such nodes can be easily found by computing a Minimum Spanning Tree (MST) over the training data, and the connected samples with different labels are then selected as prototypes. The assumption of zero error~\cite{AlleneIVC:10} holds when all arc-weights in the training set are different from each other, and its rationale is related to the fact that optimum-paths must follow the shape of the MST. However, since the ``bridges"\ between samples from different classes are protected by the prototypes (i.e., we have positive arc-weights only), there is no other way from an optimum-path to cross such a bridge (unless you have others, but then you may have more than one edge with the same weight).

In a nutshell, the training step consists in finding the \emph{optimum set} of prototypes  ${\cal P}^\ast$ and an OPF classifier rooted at ${\cal P}^\ast$. The optimum set of prototypes is defined as the one that minimizes the cost of each training sample, i.e., the training step ends up assigning an optimum cost $C(\bm{u})$ to each sample $\bm{u} \in {\cal V}$, where ${\cal V}$ stands for the training set. Such cost is computed as follows:

\begin{equation}
C(\bm{u})  =  \underset{\forall \bm{q} \in {\cal V}}{min}\{\max\{C(\bm{q}),d(\bm{q},\bm{u})\}\}, 
    \label{e.costFunction}
\end{equation}
where $C(\bm{q})=0$, $\forall q\in{\cal P}$. On the other hand, $C(\bm{q})=+\infty$, $\forall q\in{\cal V}\backslash{\cal P}$. Such costs are used to initialize the algorithm at the very first iteration.

Regarding the classification step, the model finds the optimum path from any test sample $\bm{v}$ to a prototype in ${\cal P}^\ast$ and propagates to $\bm{v}$ the prototype's label. Algorithm~\ref{a.opf_sup} implements the OPF training step.

\IncMargin{1em}
\begin{algorithm}
\SetKwInput{KwInput}{Input}
\SetKwInput{KwOutput}{Output}
\SetKwInput{KwAuxiliary}{Auxiliary}

\caption{OPF Classifier algorithm}
\label{a.opf_sup}
\Indm
\KwInput{A training set ${\cal V}$, set of prototypes ${\cal P} \subseteq {\cal V}$, map of training set labels $\lambda$,  and distance function $d$.}
\KwOutput{Predecessor map $O$, path-cost map $C$, and label map $L$.}
\KwAuxiliary{Priority queue $Q$, and variable $cst$.}
\Indp\Indpp

\BlankLine
\For{all $\bm{q} \in {\cal V}$}{
	$O(\bm{q}) \leftarrow nil$, $C(\bm{q}) \leftarrow +\infty$\;
}

\For{all $\bm{q} \in {\cal P}$}{
	$C(\bm{q}) \leftarrow 0$, $L(\bm{q}) = \lambda(\bm{q})$, $Q \leftarrow \bm{q}$\;
}

\While{$Q \neq \emptyset$}{
	Remove from $Q$ a sample $\bm{q}$ such that $C(\bm{q})$ is minimum\;

	\For {each sample $\bm{u} \in {\cal V}$ such that $\bm{q} \neq \bm{u}$ and $C(\bm{u}) > C(\bm{q})$}{
		$cst \leftarrow \max\{C(\bm{q}), d(\bm{q},\bm{u})\}$\;
		\If{$cst < C(\bm{u})$}{
			\If{$C(\bm{u}) \neq +\infty$}{
				Remove $\bm{u}$ from $Q$\;
			}
			$L(\bm{u}) \leftarrow L(\bm{q})$, $O(\bm{u}) \leftarrow \bm{q}$, $C(\bm{u}) \leftarrow cst$\;
			$Q \leftarrow \bm{u}$\;
		}
	}
}
\Return{$[O, C, L]$}
\end{algorithm}
\DecMargin{1em}

Lines $1-6$ initialize the cost map by assigning cost $0$ to the prototypes and $+\infty$ to the remaining samples. All samples have their predecessors set to $nil$, and the prototypes are inserted into the priority queue $Q$. The loop starting at Line $4$ iterates over all samples $\bm{q} \in {\cal P}$ that try to conquer the remaining nodes, and the main loop in Lines $7-19$ is the core of the OPF algorithm. A sample with minimum cost is removed from $Q$ and its neighborhood is analyzed: if the offered cost $cst$ is lower than the current cost of the prize-node $\bm{u}$, the sample $\bm{u}$ is labeled with the same label as sample's $\bm{q}$ and added to its tree (Line $15$). Notice that the classes can be represented by multiple optimum-path trees, and there must be at least one per class.

Iwashita et al.~\cite{iwashita2014path} showed how to explore some theoretical properties and links between spanning forests and the $f_{max}$ path-cost function discussed in All\'ene et al.~\cite{AlleneIVC:10} to turn the OPF training phase faster. Roughly speaking, the main idea is that the competition process ruled by supervised OPF using $f_{max}$ happens to follow the shape of the MST only, i.e., the final optimum-path forest generated after training is similar to compute the MST, then removing the arcs between prototypes, and further propagating back the costs from each prototype. Besides, it is well known that we have one possible MST only when all arc-weights are different to each other, although it may not occur in practice. In this case, there is only one path for the competition process. Since the prototypes have zero cost and all arc-weights are strictly positive, there is no other way from a prototype to conquer a sample that does not belong to its very same class.

\subsection{Unsupervised Optimum-Path Forest}
\label{ss.unsupervisedOPF}

Similarly to the supervised version, the unsupervised Optimum-Path Forest encodes each dataset sample as a node in a graph where the edges connecting a given sample to its $k$-nearest nodes are weighted according to some distance between them. Additionally, each node is weighted by a probability density function (pdf), as follows:

\begin{eqnarray}
  \rho(\bm{q}) & = & \frac{1}{\sqrt{2\pi\psi^2}k} \sum_{\forall u\in {\cal A}_k(\bm{q})} \exp\left(\frac{-d^2(\bm{q},\bm{u})}{2\psi^2}\right), \label{e.density}
\end{eqnarray}
where ${\cal A}_k(\bm{q})$ stands for the $k$-neighboorhod of sample $\bm{q}$, $\psi = \frac{d_f}{3}$, and $d_f$ is the maximum weight among the edges in the graph $({\cal V},{\cal A}_k)$. 

The traditional method to estimate the probability density is through a Parzen-window. However, such an approach requires finding the optimum number of nearest neighbors $k^{\ast} \in \{1 \leq  k_{\max}\leq |{\cal V}|\}$, where $k_{max}$ is an ad-hoc parameter. To tackle such an issue, Rocha et al.~\cite{RochaIJIST:09} proposed to find the $k^{\ast}$ that minimizes the graph cut over ${\cal V}$ instead.

The unsupervised OPF defines the set of prototype nodes ${\cal P}$ composed of one element per maximum of the pdf. A path $\phi_{\bm{u}}$ is considered optimum if, given a path value $f_{min}(\phi_{\bm{u}})$, $f_{min}(\phi_{\bm{u}})\geq f_{min}(\tau_{\bm{u}})$ for any other path $\tau_{\bm{u}}$. Finally, $\bm{u}$ is assigned to the path whose minimum density value along it is maximum, defined as follows:

\begin{eqnarray}
\label{e.pf2}
f_{min}(\langle u \rangle) & = & \left\{ \begin{array}{ll} 
    \rho(\bm{u})           & \mbox{if $\bm{u} \in {\cal P}$} \\
    \rho(\bm{u}) - \delta  & \mbox{otherwise,}
 \end{array}\right. \nonumber \\
f_{min}(\langle \phi_q\cdot \langle \bm{q},\bm{u}\rangle\rangle)&=& \min \{f_{min}(\phi_{\bm{q}}), \rho(\bm{u})\},
\end{eqnarray}
where $\delta$ is small constant. Algorithm~\ref{a.unsupopf} implements the unsupervised OPF.

\IncMargin{1em}
\begin{algorithm}
\SetKwInput{KwInput}{Input}
\SetKwInput{KwOutput}{Output}
\SetKwInput{KwAuxiliary}{Auxiliary}

\caption{OPF Clustering algorithm}
\label{a.unsupopf}
\Indm
\KwInput{Graph $({\cal V},{\cal A}_k)$.}
\KwOutput{Predecessor map $O$, path-cost map $C$, and label map $L$.}
\KwAuxiliary{Priority queue $Q$, density map $\rho$, variables $cst$, and $l\leftarrow 1$.}
\Indp\Indpp

\BlankLine
\For{all $\bm{q} \in {\cal V}$}{
	Compute $\rho(q)$ using Equation~\ref{e.density}\;
	$O(\bm{q})\leftarrow nil$, $C(\bm{q})\leftarrow \rho(\bm{q})-\delta$, $Q\leftarrow \bm{q}$\;
}

\While{$Q \neq \emptyset$}{
	Remove from $Q$ a sample $\bm{q}$ such that $C(\bm{q})$ is maximum\;
	\If{$O(\bm{q}) = nil$}{
		$L(\bm{q})\leftarrow l$, $l\leftarrow l + 1$, and $C(\bm{q})\leftarrow \rho(\bm{q})$\;
	}
	\For {all $\bm{u}\in {\cal A}_k(\bm{q})$ such that $C(\bm{u}) < C(\bm{q})$}{
		$cst \leftarrow \min\{C(\bm{q}), \rho(\bm{u})\}$\;

		\If{$P(tmp > C(\bm{u})$}{
			$L(\bm{u})\leftarrow L(\bm{q})$, $P(\bm{u})\leftarrow \bm{q}$, $C(\bm{u})\leftarrow tmp$\;
			Update position of $\bm{u}$ in $Q$\;
		}
	}
}
\Return $[P,C,L]$
\end{algorithm}
\DecMargin{1em}

In \emph{Algorithm~\ref{a.unsupopf}}, Lines $1-4$ initialize the variables, and also inserts all samples in the priority queue $Q$. The main loop in Lines $5-17$ is responsible for the unsupervised OPF algorithm. It first removes a sample $\bm{q}$ from $Q$ with maximum path value $C(\bm{q})$ in Line $6$. If $\bm{q}$ has not been conquered by any other sample, then $O(\bm{q})=nil$ (Line $7$) and $\bm{q}$ is a prototype of the connectivity map (a maximum of the pdf). Since $q\in {\cal P}$, by Equation~(\ref{e.pf2}), its connectivity value is reset to $\rho(\bm{q})$ (Line 8), which in addition to the fact that ${\cal A}_k$ is symmetric on plateaus of the pdf, will make root $\bm{q}$ to conquer the remaining samples of its plateau. It is also assigned to it a new distinct label (cluster) for optimum-path propagation to the rest of its dome. The inner loop in Lines $10-16$ evaluates all adjacent sample $u$ of $\bm{q}$ to which $\bm{q}$ can offer a better connectivity value (i.e., $C(\bm{u}) < C(\bm{q})$). If the path $\phi_{\bm{q}}\cdot \langle \bm{q},\bm{u}\rangle$ offers a higher cost to $\bm{u}$ (Lines $11-12$), then the current path $\phi_{\bm{u}}$ is substituted by the new path $\phi_{\bm{q}}\cdot \langle \bm{q},\bm{u}\rangle$, being the maps $C(\bm{u})$, $L(\bm{u})$, and $O(\bm{u})$ updated accordingly (Lines $13-14$).

\section{Fuzzy Optimum-Path Forest}
\label{s.fopf}

In this section, we present the proposed Fuzzy Optimum-Path Forest (Fuzzy OPF), a new method for data classification that extends na\"{i}ve OPF classifier to fuzzy operators. The proposed approach is divided into two main steps: (i) to compute each sample's membership, which is performed in a non-supervised fashion through an adapted version of the OPF algorithm, and (ii) to introduce such an information in the path-cost function formulation.

\subsection{Membership Computation}
\label{ss.fuzzyMembership}

Although OPF has presented itself as an interesting method for classification problems~\cite{iwashita2014path,rebouccas2017automated}, it lacks real-world applications where samples may belong partially to a given class. Suppose a sample tagged with a given label despite its features diverge considerably from the other samples with the same label. Such an outlier contributes equally to the training step of standard OPF, i.e., it has the same importance, although acting like a ``noise" and leading the learning process to overfit~\cite{PASSOSEPSR:2016}.

\begin{figure}[!h]
  \centerline{
    \begin{tabular}{c}
	\includegraphics[width=7.5cm]{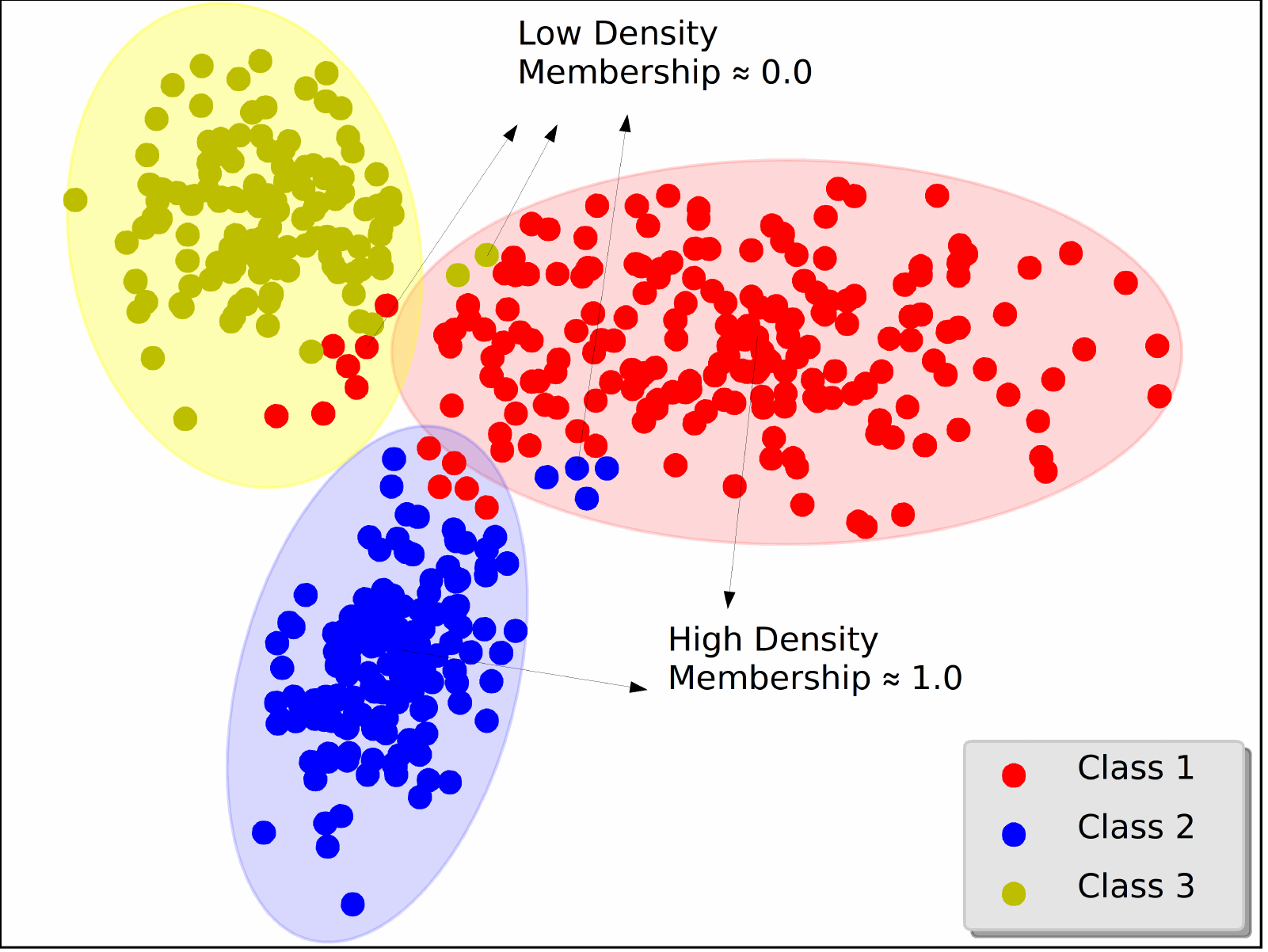} \\
    \end{tabular}}	
  \caption{\label{f.fuzzyMembership} membership computation process. Samples located near to the cluster center presents, in general, higher densities. On the other hand, samples away from their cluster center tend to bear low densities, which implies in small values of the membership. Such process penalizes distant samples, thus helping handling the overfitting problem.}
\end{figure}

In this work, we consider attributing a membership $F_\Theta(\bm{x})\in[0,1]$ to each training sample ${\bm x}$, i.e., a measure of how meaningful is a sample regarding its class, where $\Theta$ stands for the set of function parameters. Such a function describes the influence of sample $\bm{x}$ in its group. An adequate choice of the membership lies over two constraints: (i) selecting a proper lower bound parameter $\sigma > 0$, which is performed using a grid search, and (ii) defining the model properties that best describe the data behavior~\cite{lin2002fuzzy}. Such information can be encoded by the node density, which is computed in an unsupervised fashion using the OPF technique. The membership is calculated as follows:

\begin{equation}
    \label{e.fuzzyMembership}    
    F_\Theta(\bm{x}) = (1-\sigma) \bigg(\dfrac{\rho(\bm{x})-\rho_{min}}{\rho_{max}-\rho_{min}}\bigg)^2 + \sigma,
\end{equation}
where $\rho_{min} \leq  \rho(\bm{x}) \leq \rho_{max}$, and $\rho_{min}$ and $\rho_{max}$ stand for the lowest and the highest densities, respectively, and $\Theta=\{\sigma,\rho_{min},\rho_{max}\}$. Figure~\ref{f.fuzzyMembership} illustrates the concept behind such an idea, where samples located at the boundaries of the clusters tend to possess smaller membership values, i.e., they are supposed to be ``less reliable"\ during the competition process.

\subsection{Fuzzy Optimum-Path Forest}
\label{ss.fuzzyOPF}

Since the OPF training step is described as a competition process among prototypes to ``conquer'' other samples, we introduce the membership in the cost formulation as follows:

\begin{equation}
C(\bm{u})  =  \underset{\forall \bm{q} \in {\cal V}}{min}\{F_\Theta(\bm{u}) *\max\{C(\bm{q}),d(\bm{q},\bm{u})\}\}, 
\label{e.pathCostFuzzy}
\end{equation}
which is basically a reformulation of the supervised OPF cost function (Equation~\ref{e.costFunction}). Notice the above equation is employed during the test phase only since the membership values are computed during training 

Smaller membership values stand for examples that have no importance for training purposes, being often called anomalies or outliers. On the other hand, high membership values indicate the most representative samples, with the maximum values attributed to the prototypes. Finally, values within that boundaries stand for the transition between these two states. 

Let us suppose that $\sigma = 0$ and $\rho_{min} = \rho(\bm{x})$ for some sample $\bm{x}\in{\cal V}$. Therefore, we have that  $F_\theta(\bm{x})\approx 0$ in Equation~\ref{e.fuzzyMembership}. In such a case, $C(\bm{x})$ is always equals to $0$ in Equation~\ref{e.pathCostFuzzy}, i.e., the OPF properties are no longer held or erratic. A sample with the lowest pdf value, i.e., $\rho_{min}$, means that it has been placed in a less dense region of the feature space. In this case, either $\bm{x}$ may figure as an outlier or a less representative sample. However, in this paper, we limited the values of sigma within the range $[0.2,1.2]$.

Algorithm~\ref{a.fuzzyOPF} implements the Fuzzy OPF classifier. First, Lines $1-3$ compute each sample's density, and Line $4$ sets $\rho_{min}$ and $\rho_{max}$ to the minimum and maximum densities, respectively. Such values are used to compute each sample's membership, as presented in Line $7$. Lines $9-11$ initialize the prototypes with zero cost, its true label (i.e., function $\lambda$), and inserts them in the priority queue.

Afterward, the membership (Equation~\ref{e.pathCostFuzzy}) is employed in line $15$ to calculate the cost of each sample, The remainder of the algorithm follows the same idea presented in Algorithm~\ref{a.opf_sup}, i.e., the main loop in lines $12-25$ are in charge of the competition process among prototype samples, and the algorithm outputs the cost map $C$, predicted labels in $L$, and the optimum-path forest stored in $O$.

\IncMargin{1em}
\begin{algorithm}
\SetKwInput{KwInput}{Input}
\SetKwInput{KwOutput}{Output}
\SetKwInput{KwAuxiliary}{Auxiliary}

\caption{Fuzzy OPF Classifier algorithm}
\label{a.fuzzyOPF}
\Indm
\KwInput{Graph $({\cal V},{\cal A}_k)$, set of prototypes ${\cal P} \subseteq {\cal V}$, map of training set labels $\lambda$, lower bound parameter $\sigma$, and distance function $d$.}
\KwOutput{Predecessor map $O$, path-cost map $C$, and label map $L$.}
\KwAuxiliary{Priority queue $Q$, variable $cst$, density map $\rho$, and minimum and maximum densities $\rho_{min}$ and $\rho_{max}$, respectively.}
\Indp\Indpp

\BlankLine
\For{all $\bm{q} \in {\cal V}$}{
	Compute $\rho(\bm{q})$ using Equation~\ref{e.density}\;
}
$\rho_{min}, \rho_{max}\leftarrow min(\rho), max(\rho)$\;

\For{all $\bm{q} \in {\cal V}$}{
	$O(\bm{q}) \leftarrow nil$, $C(\bm{q}) \leftarrow +\infty$\;
	Compute $F_\Theta(\bm{q})$ using Equation~\ref{e.fuzzyMembership}\;
}

\For{all $\bm{q} \in {\cal P}$}{
	$C(\bm{q}) \leftarrow 0$, $L(\bm{q}) = \lambda(\bm{q})$, $Q \leftarrow \bm{q}$
}

\While{$Q \neq \emptyset$}{
	Remove from $Q$ a sample $\bm{q}$ such that $C(\bm{q})$ is minimum\;

	\For {each sample $\bm{u} \in {\cal V}$ such that $\bm{q} \neq \bm{u}$ and $C(\bm{u}) > C(\bm{q})$}{
		$cst \leftarrow F_\Theta(\bm{u}) * \max\{C(\bm{q}), d(\bm{q},\bm{u})\}$\;
		\If{$cst < C(\bm{u})$}{
			\If{$C(\bm{u}) \neq +\infty$}{	Remove $\bm{u}$ from $Q$}
			$L(\bm{u}) \leftarrow L(\bm{q})$, $O(\bm{u}) \leftarrow \bm{q}$, $C(\bm{u}) \leftarrow cst$\;
			$Q \leftarrow \bm{u}$\;
		}
	}
}
\Return{$[O, C, L]$}
\end{algorithm}
\DecMargin{1em}

The loop in Lines $9-11$ is in charge of assigning the true label map ($\lambda$) to all prototypes (${\cal P}$). According to the OPF working mechanism, the prototypes are the first samples that will be removed from the priority queue ${\cal Q}$ in Line $13$ since they have zero cost (Line $10$). This means their true labels will be propagated first to the samples they have conquered (Line $20$), and later on to the next samples from ${\cal Q}$.

\subsection{Computational Complexity}
\label{ss.computationalComplexity}

Regarding the computational complexity, let us analyze standard OPF first. From Algorithm~\ref{a.opf_sup}, Lines $7-19$ stand for the core of OPF technique. Since we have $\left|{\cal V}\right|$ training samples added to the priority queue $Q$ and each sample is removed only once, the main loop runs $\theta(\left|{\cal V}\right|)$ times. Besides, since we used a binary heap to implement the priority queue, Line $8$ requires $O(\log\left|{\cal V}\right|)$ computations. However, since we have a complete graph, the inner loop in Lines $8-18$ executes $\theta(\left|{\cal V}\right|)$ times. The final complexity for training is $\theta(\left|{\cal V}\right|^2)$. The classification step can be performed in $\theta(\left|{\cal V}\right|.\left|{\cal T}\right|)$, where $\left|{\cal T}\right|$ stands for the test set. However, Papa et al.~\cite{PapaPR:2012} that such an step can be optimized by storing the training samples in an ascending order of costs. Additionally, Iwashita et al.~\cite{iwashita2014path} highlighted how to make OPF training step faster by exploring some theoretical properties presented by All\'ene et al.~\cite{AlleneIVC:10}.

Regarding Fuzzy-OPF, we must analyze the complexity of OPF clustering first, which is composed of two main steps: (i) to find $k^\ast$ and (ii) to execute the competition process (Algorithm~\ref{a.unsupopf}). With respect to the first step, for each value $k\in[1,k_{max}]$, we need to find the $k$-neighborhood and compute the minimum graph cut. The former can be computed in $\theta(\left|{\cal V}\right|^2)$ using any standard data structure (i.e., arrays), and the second step takes $\theta(k\left|{\cal V}\right|)$ iterations. The whole procedure than takes $k_{max}(\theta(\left|{\cal V}\right|^2)+\theta(k\left|{\cal V}\right|))$. If $k_{max}\rightarrow \left|{\cal V}\right|$, then the complexity is done by $\left|{\cal V}\right|.\theta(\left|{\cal V}\right|^2+\left|{\cal V}\right|^2)=\left|{\cal V}\right|.\theta(2\left|{\cal V}\right|^2)\in\left|{\cal V}\right|.\theta(\left|{\cal V}\right|^2)=\theta(\left|{\cal V}\right|^3)$ since $k\in\theta(k_{max})$. However, we limited $k_{max}$ to $100$ in the paper since we observed that greater values did not contribute to the final results. In practice, the final complexity of OPF clustering is $\theta(\left|{\cal V}\right|^2)$ in this paper because $\left|{\cal V}\right|>>100$.

Concerning the second step, the loop in Lines $1-4$ executes $\theta(\left|{\cal V}\right|)$ times, while Line $2$ runs over all $k$ neighbors of each sample $\bm{q}$. Therefore, the complexity of lines $1-4$ in Algorithm~\ref{a.unsupopf} is represented by $\theta(k^\ast\left|{\cal V}\right|)$. The competition process is performed in Lines $5-17$, which runs over all training samples in $Q$, i.e., it requires $\theta(\left|{\cal V}\right|)$ iterations. Line $6$ runs in $O(\log\left|{\cal V}\right|)$ due to the binary heap, and the inner loop in Lines $10-16$ requires $k^\ast$ iterations. Therefore, the complexity for training OPF clustering is $\theta(k^\ast\left|{\cal V}\right|\log\left|{\cal V}\right|)$. Notice that when $k^\ast\rightarrow\left|{\cal V}\right|$, the complexity is done by $\theta(\left|{\cal V}\right|^2)$. Therefore, the complexity of Fuzzy-OPF requires $\theta(\left|{\cal V}\right|^2)+\theta(\left|{\cal V}\right|^2)\in\theta(\left|{\cal V}\right|^2)$ operations.

\section{Methodology}
\label{s.methodology}

In this section, we introduce the datasets and the experimental setup employed in the work.

\subsection{Dataset Description}
\label{ss.datasets}

The proposed approach is evaluated over twelve datasets for general-purpose classification problems. Six datasets were synthetically generated for testing purposes, and the remaining ones stand for real-world problems. Among these, four stand for public datasets and two concern private data, as described below:

\subsubsection{Synthetic Datasets}
\label{sss.synthetic_datasets}

\begin{figure}[!h]
  \centerline{
   \begin{tabular}{cc}
	\includegraphics[width=4.5cm]{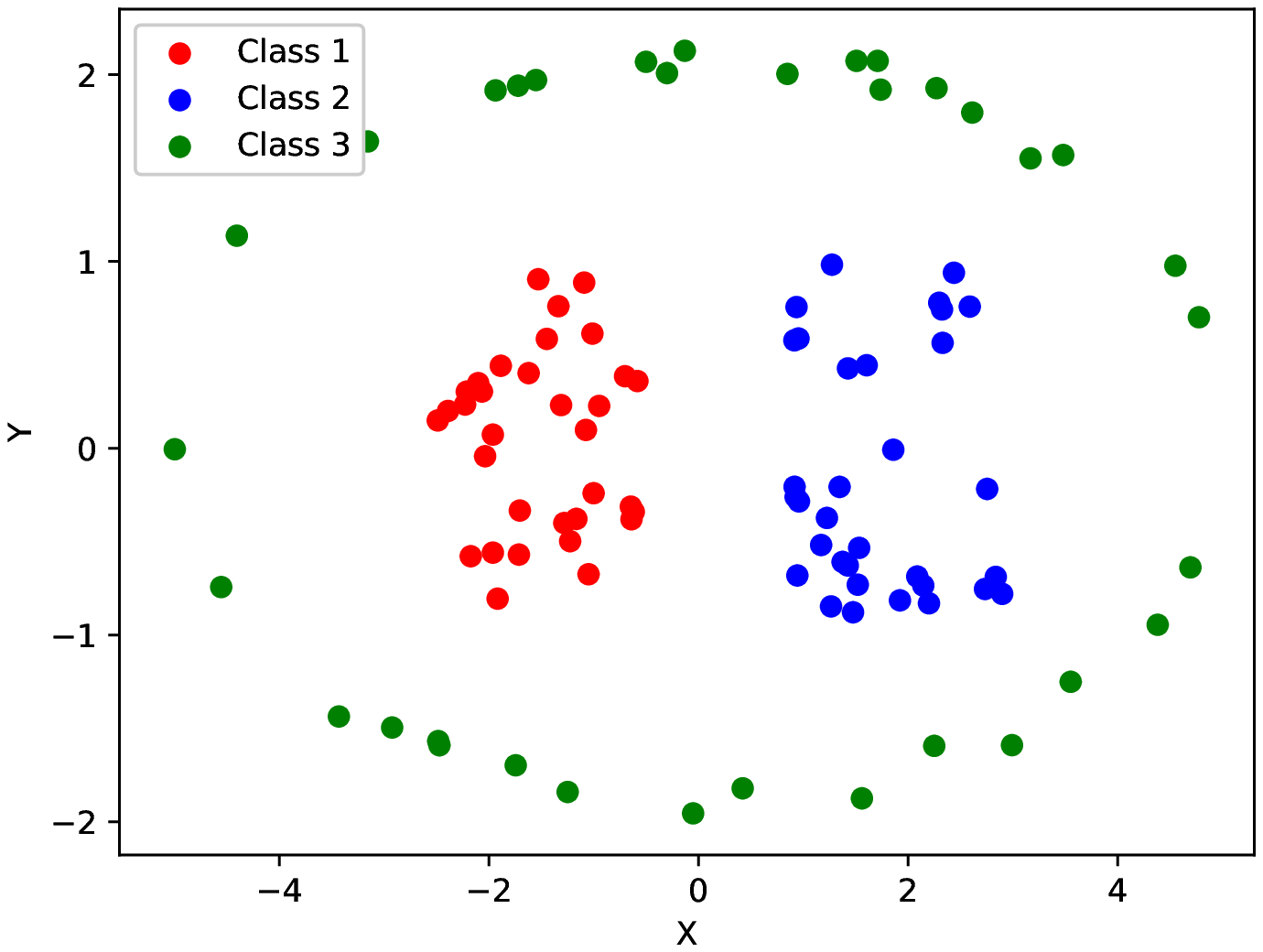} &
	\includegraphics[width=4.5cm]{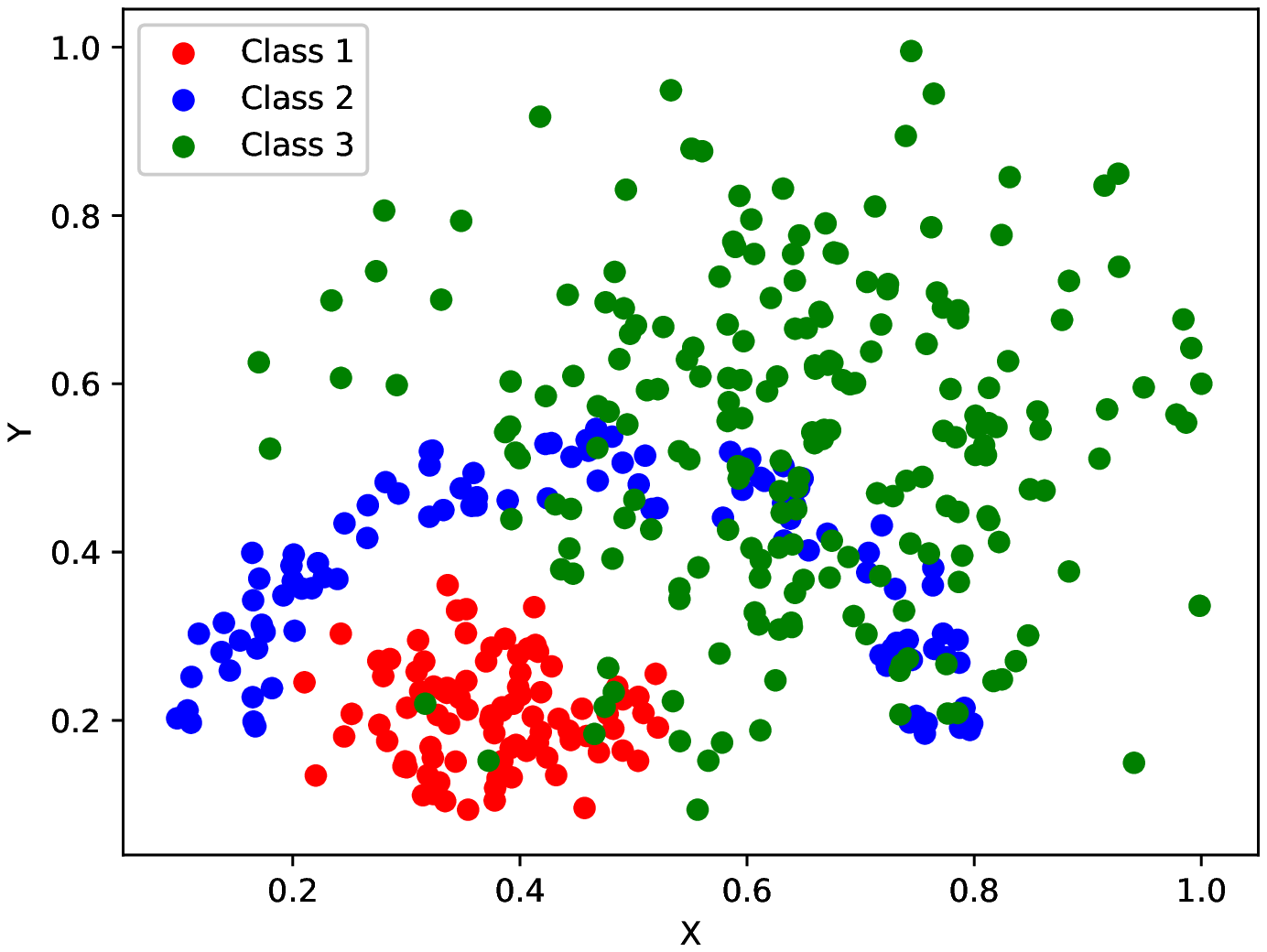} \\
	(a) & (b)
   \end{tabular}}	
   \centerline{
   \begin{tabular}{cc}
	\includegraphics[width=4.5cm]{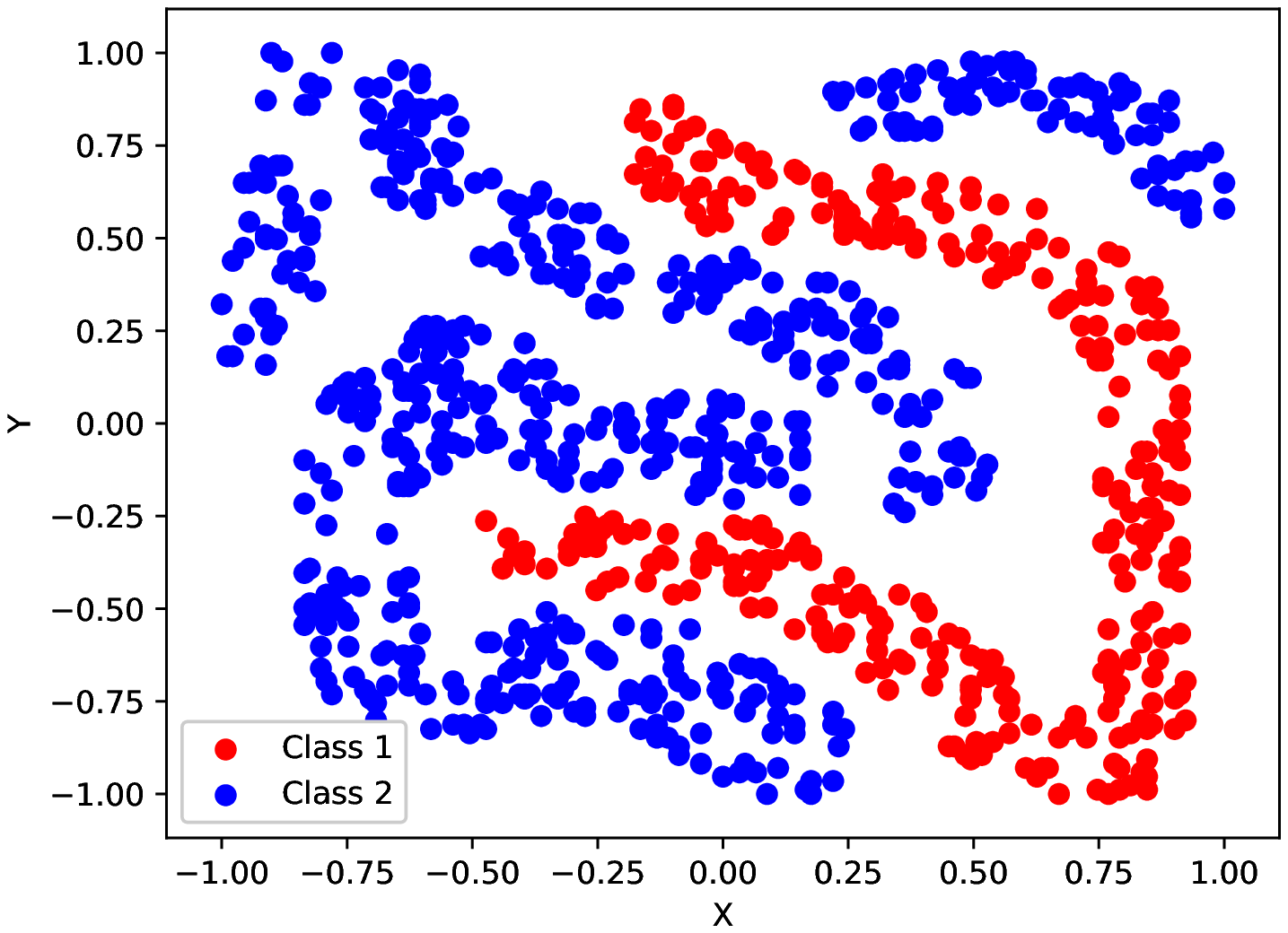} &
	\includegraphics[width=4.5cm]{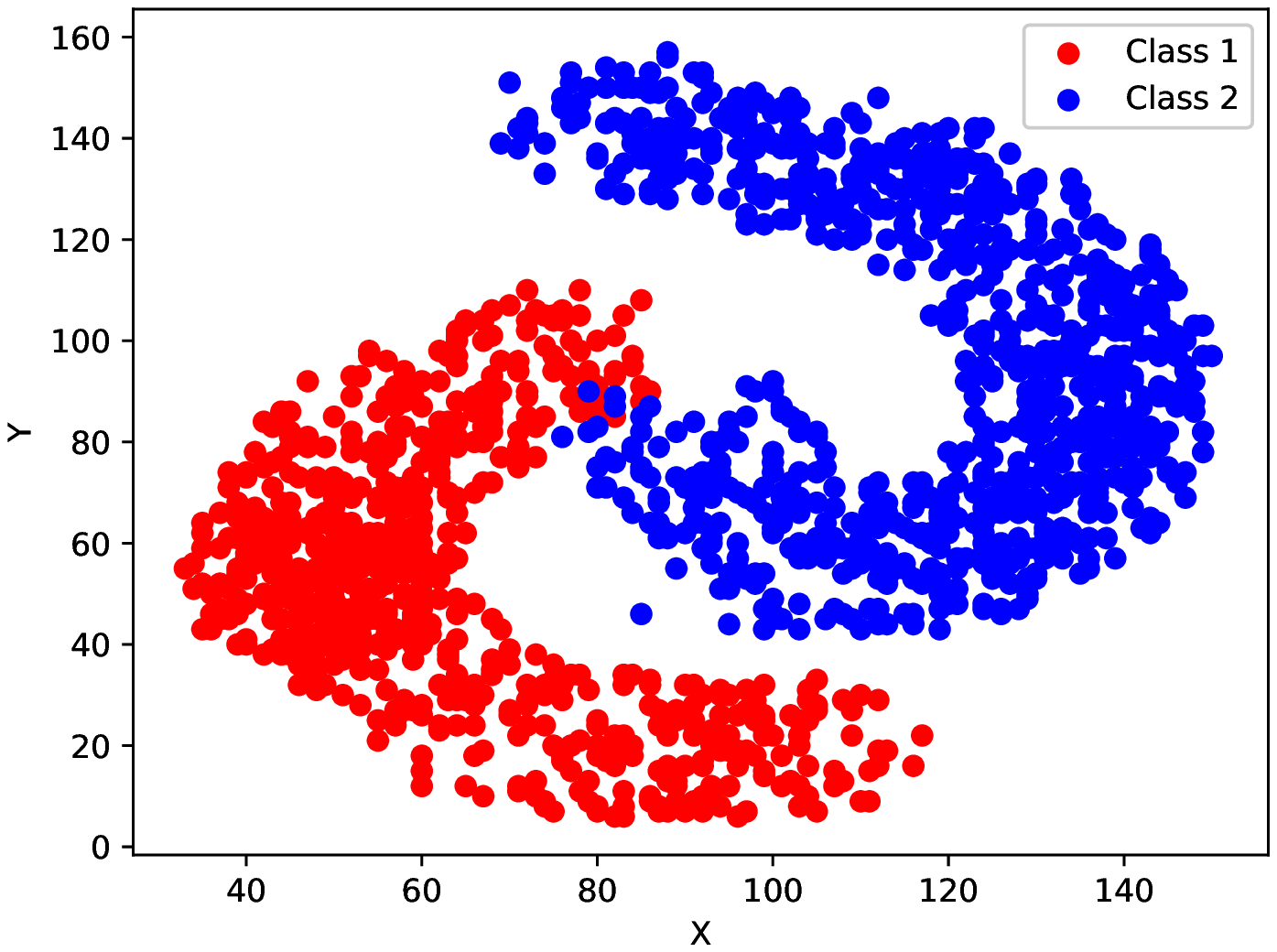}\\
	(c) & (d)
   \end{tabular}}
   \centerline{
   \begin{tabular}{cc}
	\includegraphics[width=4.5cm]{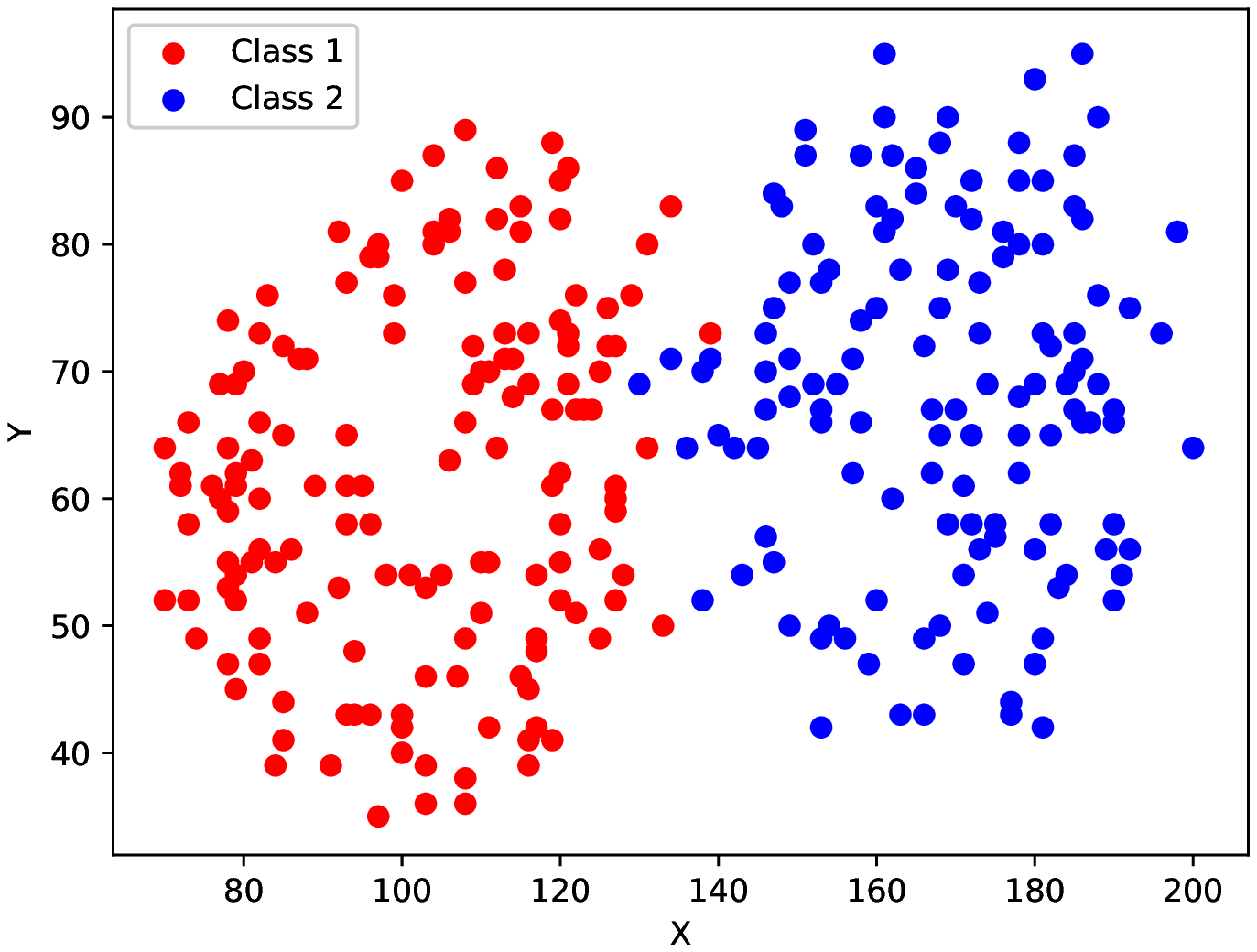} &
	\includegraphics[width=4.5cm]{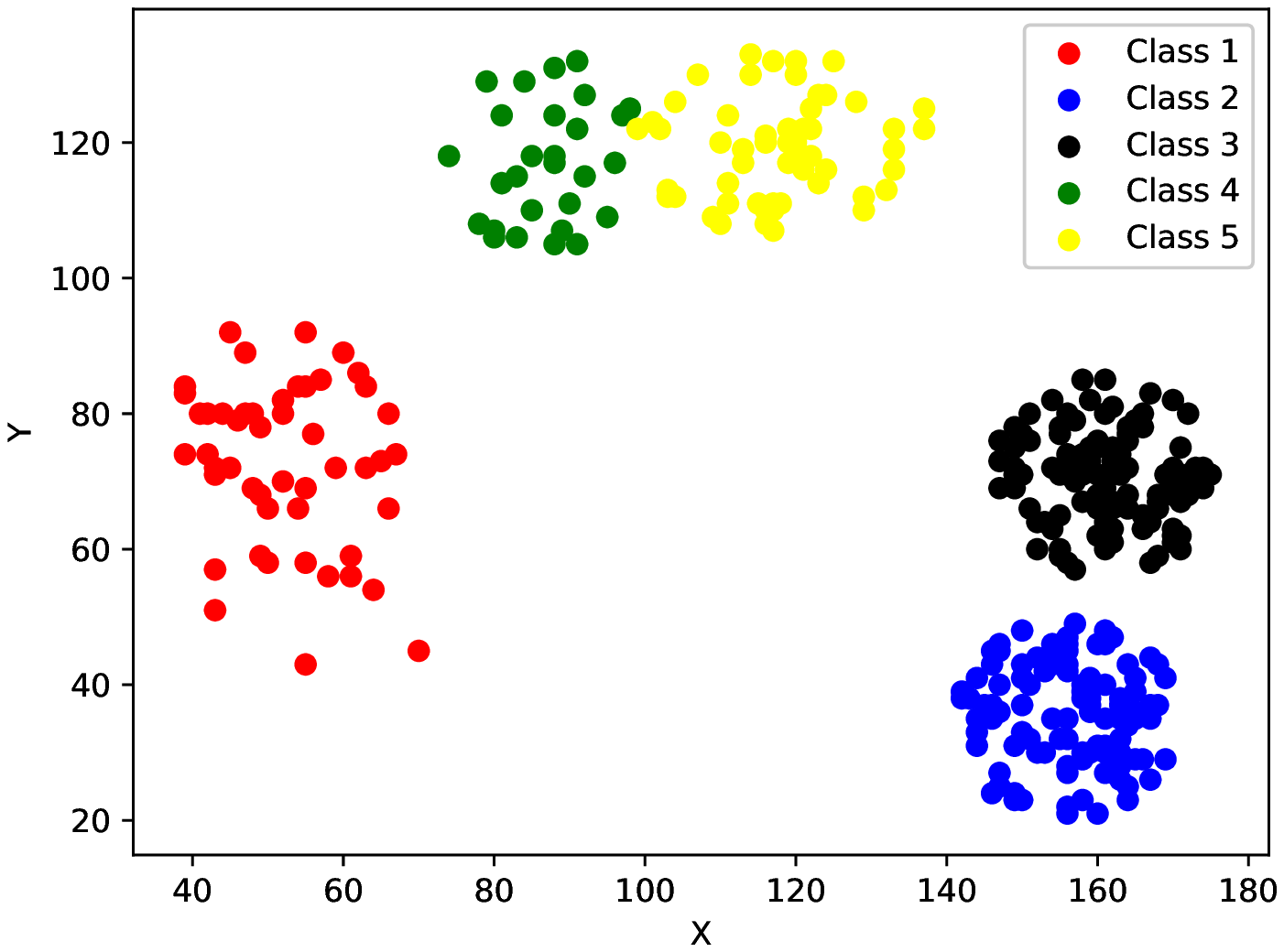}\\
	(e) & (f)
    \end{tabular}}
  \caption{\label{f.kuncheva} Datasets: (a) Boat, (b) Cone-Torus, (c) Four-Class, (d) Data 1, (e) Data 2, and (f) Data 3.}
\end{figure}

\begin{itemize}
\item Boat: dataset containing $100$ samples represented by two features and distributed into $3$ classes\footnote{\url{https://github.com/jppbsi/LibOPF/tree/master/data}}. 
\item Cone-Torus: dataset containing $400$ samples represented by two features and distributed into $3$ classes$^2$. 
\item Four-Class: dataset containing $100$ samples represented by two features and distributed into $2$ classes\footnote{\url{https://www.csie.ntu.edu.tw/~cjlin/libsvmtools/datasets/binary.html}}.
\item Data1: dataset containing $1,423$ samples represented by two features and distributed into $2$ classes$^2$. 
\item Data2: dataset containing $283$ samples represented by two features and distributed into $2$ classes$^2$. 
\item Data3: a dataset containing $340$ samples represented two features and distributed into $5$ classes$^2$.    
\end{itemize}
Figure~\ref{f.kuncheva} depicts the above datasets.   

\subsubsection{Real Datasets}
\label{sss.real_datasets}

\begin{figure}[!h]
  \centerline{
    \begin{tabular}{c}
	\includegraphics[width=4.5cm]{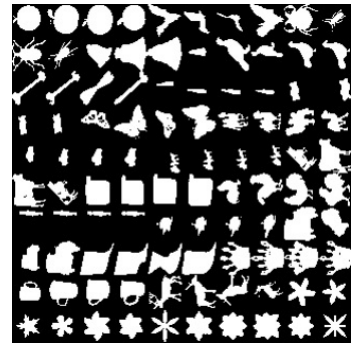} \\
    \end{tabular}}	
  \caption{\label{f.mpeg7} Some MPEG-$7$ image samples.}
\end{figure}

\begin{itemize}
\item Thyroid: dataset containing $7,200$ samples represented by a $21$-dimensional vector each and distributed into $2$ classes\footnote{\url{http://archive.ics.uci.edu/ml/datasets}}. 
\item Breast Tissue: dataset containing $106$ samples represented by a $10$-dimensional vector each and distributed into $6$ classes$^{4}$.
\item Landsat Satellite: dataset containing $5,100$ samples represented by a $36$-dimensional vector each and distributed into $8$ classes\footnote{\url{https://archive.ics.uci.edu/ml/datasets/Statlog+(Landsat+Satellite)}}. 
\item MPEG-$7$ BAS: The MPEG-$7$~\cite{MPEG-7} shape dataset contains $1,400$ samples distributed into $70$ classes. This work employed BAS (Beam Angle Statistics)~\cite{AricaPRL:03} shape descriptor to extract $180$ features from the whole dataset. Details about the features extraction procedure can be found in~\cite{PapaIJIST09}. Figure~\ref{f.mpeg7} depicts some MPEG-$7$ sample images$^2$.
\item Electric Industrial Profiles: private dataset containing $3,178$ samples represented by an $8$-dimensional vector each and distributed into $2$ classes~\cite{PASSOSEPSR:2016}.
\item Electric Commercial Profiles: private dataset containing $4,952$ samples represented by an $8$-dimensional vector each and distributed into $2$ classes~\cite{PASSOSEPSR:2016}.
\end{itemize}

\subsection{Experimental Setup}
\label{ss.experimentalSetup}

The experiments compared the Fuzzy OPF against three supervised learning algorithms: naive OPF, Linear SVM, and the well-known Bayesian classifier. The methodology employed a cross-validation procedure with $20$ runs to provide a statistical analysis using the Wilcoxon signed-rank test with a significance of $0.05$~\cite{Wilcoxon:45}. This test repeatedly measures data dependency and computes the difference of samples when they are matched. These differences are ranked for further applying a negative sign to all the ranks where the difference between two samples is negative, denoting the signed-rank. Such an approach is robust against outliers and heavy tail distributions, also presenting itself as a good test to compare mean scores when the dependent variable is not normally distributed. For each run, the dataset is randomly split into $60\%$ for training, $20\%$ for evaluation, and $20\%$ for testing purposes. 

The evaluation set is used to fine-tune hyperparameter $k_{max}$ and $\sigma$  through a grid-search within the intervals: $k_{max}\in\{1,10,20,\ldots,150\}$ and $\sigma\in\{0.2,0.4,0.6,0.8,1.0,1.2\}$. Then, we take the values that maximize the Fuzzy OPF accuracy. The same approach was conducted towards SVM hyperparameter. 

Finally, the experiments were developed over the C-based library LibOPF\footnote{https://github.com/jppbsi/LibOPF}, which implements the Optimum-Path Forest framework. Additionally, we used SVM and Bayes implementations provided by Scikit-learn~\cite{scikit-learn}. Regarding the computational environment, we employed an $1.70$ GHz $2$x Intel$^{\textregistered}$ Xeon Bronze $3106$ processor with 64GB of RAM, running over an Ubuntu 16.04 Linux machine.

\section{Results and Discussions}
\label{s.experiments}

In this section, we discuss the experimental results concerning the proposed approach, as well as the procedure to fine-tune the hyperparameters.

\subsection{General-Purpose Datasets Classification}
\label{ss.testing}

Table~\ref{t.results} presents the mean accuracy and its standard deviation, as well as the mean F$1$-measure values. Notice the best values according to the Wilcoxon signed-rank test are highlighted in bold.

\begin{table}[!htb]
\caption{Average accuracy and its standard deviation, as well as the average F$1$-measure considering the test set.}
\begin{center}
\renewcommand{\arraystretch}{1.5}
\setlength{\tabcolsep}{6pt}
\resizebox{\columnwidth}{!}{
\begin{tabular}{l|c|c|c|c|c}
\hhline{-|-|-|-|-|-|}
\hhline{-|-|-|-|-|-|}
\hhline{-|-|-|-|-|-|}
\cline{3-6}
\multicolumn{1}{c|}{\cellcolor[HTML]{EFEFEF}{\textbf{Dataset}}}  & \multicolumn{1}{c|}{\cellcolor[HTML]{EFEFEF}{\textbf{Statistics}}}   & \cellcolor[HTML]{EFEFEF}{\textbf{OPF Classifier}} & \cellcolor[HTML]{EFEFEF}{\textbf{Fuzzy-OPF}} & \cellcolor[HTML]{EFEFEF}{\textbf{SVM}} & \cellcolor[HTML]{EFEFEF}{\textbf{Bayes}}\\ \hline
\multicolumn{1}{c|}{Boat} & Mean Acc. & \textbf{0.99286} & \textbf{0.99286} & 0.70240 & \textbf{0.98096}\\
\multicolumn{1}{c|}{ } & Std Acc. & 0.01700 & 0.01700 & 0.03648 & 0.04617\\
\multicolumn{1}{c|}{ } & Mean F1 & 0.99282 & 0.99282 & 0.62437 & 0.97998\\ \hline
\multicolumn{1}{c|}{Cone-Torus} & Mean Acc. & 0.82717 & \textbf{0.83458} & 0.75245 & \textbf{0.81607}\\
\multicolumn{1}{c|}{ } & Std Acc. & 0.02974 & 0.03109 & 0.04459 & 0.03512\\
\multicolumn{1}{c|}{ } & Mean F1 & 0.82896 & 0.83603 & 0.71438 & 0.80278\\ \hline
\multicolumn{1}{c|}{Four-Class} & Mean Acc. & \textbf{0.99797} & \textbf{0.99884} & 0.76560 & 0.75201\\
\multicolumn{1}{c|}{ } & Std Acc. & 0.00332 & 0.00232 & 0.02509 & 0.02146\\
\multicolumn{1}{c|}{ } & Mean F1 & 0.99797 & 0.99884 & 0.75625 & 0.74916\\\hline\hline
\multicolumn{1}{c|}{Data1} & Mean Acc. & \textbf{0.99458} & \textbf{0.99475} & 0.95174 & 0.94232\\
\multicolumn{1}{c|}{ } & Std Acc. & 0.00421 & 0.00375 & 0.01061 & 0.01160\\
\multicolumn{1}{c|}{ } & Mean F1 & 0.99458 & 0.99475 & 0.95176 & 0.94233\\ \hline
\multicolumn{1}{c|}{Data2} & Mean Acc. & \textbf{0.98277} & \textbf{0.98277} & \textbf{0.98364} & \textbf{0.98621}\\
\multicolumn{1}{c|}{ } & Std Acc. & 0.01635 & 0.01635 & 0.01587 & 0.01401\\
\multicolumn{1}{c|}{ } & Mean F1 & 0.98276 & 0.98276 & 0.98362 & 0.98619\\\hline
\multicolumn{1}{c|}{Data3} & Mean Acc. & \textbf{0.99643} & \textbf{0.99643} & \textbf{0.99428} & \textbf{0.99714}\\
\multicolumn{1}{c|}{ } & Std Acc. & 0.00619 & 0.00619 & 0.00834 & 0.00572\\
\multicolumn{1}{c|}{ } & Mean F1 & 0.99635 & 0.99638 & 0.99432 & 0.99711\\\hline\hline

\multicolumn{1}{c|}{Thyroid} & Mean Acc. & 0.97140 & 0.97191 & \textbf{0.97568} & 0.13945\\
\multicolumn{1}{c|}{ } & Std Acc. & 0.00250 & 0.00258 & 0.00299 & 0.05072\\
\multicolumn{1}{c|}{ } & Mean F1 & 0.96879 & 0.96920 & 0.97161 & 0.18716\\\hline
\multicolumn{1}{c|}{Breast Tissue} & Mean Acc. & \textbf{0.67501} & \textbf{0.67917} & \textbf{0.68959} & \textbf{0.70833}\\ 
\multicolumn{1}{c|}{ } & Std Acc. & 0.06922 & 0.06333 & 0.07623 & 0.07683\\
\multicolumn{1}{c|}{ } & Mean F1 & 0.66114 & 0.66551 & 0.68329 & 0.70283\\ \hline

 \multicolumn{1}{c|}{Landsat Satellite} & Mean Acc. & \textbf{0.99244} & \textbf{0.99264} & \textbf{0.99260} & 0.93725\\
\multicolumn{1}{c|}{ } & Std Acc. & 0.00178 & 0.00187 & 0.00190 & 0.01287\\
\multicolumn{1}{c|}{ } & Mean F1 & 0.99177 & 0.99201 & 0.99200 & 0.95670\\\hline \hline

\multicolumn{1}{c|}{MPEG-7 BAS} & Mean Acc. & 0.80179 & \textbf{0.80732} & 0.77892 & 0.71465\\
\multicolumn{1}{c|}{ } & Std Acc. & 0.02026 & 0.01909 & 0.01987 & 0.02044\\
\multicolumn{1}{c|}{ } & Mean F1 & 0.78661 & 0.79210 & 0.76479 & 0.71905\\\hline

\multicolumn{1}{c|}{Elec. Ind. Prof.} & Mean Acc. & \textbf{0.94899} & \textbf{0.94938} & 0.93720 & 0.40754\\
\multicolumn{1}{c|}{ } & Std Acc. & 0.00687 & 0.00674 & 0.00000 & 0.02120\\
\multicolumn{1}{c|}{ } & Mean F1 & 0.94972 & 0.95011 & 0.90680 & 0.51753\\\hline

\multicolumn{1}{c|}{Elec. Com. Prof.} & Mean Acc. & \textbf{0.96336} & \textbf{0.96372} & 0.94550 & 0.90495\\
\multicolumn{1}{c|}{ } & Std Acc. & 0.00586 & 0.00561 & 0.00000 & 0.00942\\
\multicolumn{1}{c|}{ } & Mean F1 & 0.96375 & 0.96403 & 0.91900 & 0.90310\\

\hhline{-|-|-|-|-|-|}
\hhline{-|-|-|-|-|-|}
\hhline{-|-|-|-|-|-|}
\end{tabular}}
\label{t.results}
\end{center}
\end{table}

From Table~\ref{t.results}, one can draw several conclusions: (i) Fuzzy OPF is the only technique capable of achieving the most accurate results alone, according to the Wilcoxon signed-rank test, over MPEG-BAS dataset; (ii) Fuzzy OPF obtained the best results considering eight out of twelve datasets, i.e., Boat, Cone-Torus, Four-Class, Data1, Landsat Satellite, MPEG-BAS, Elec. Ind. Prof., and Elec. Com. Prof.; (iii) Fuzzy OPF obtained the most accurate results (i.e., statistically speaking), considering the Wilcoxon similarity test, in eleven out of twelve datasets. Moreover, it worth highlighting the Fuzzy OPF achieved, in the worst case, identical results to the standard OPF. Such behavior is expected since Fuzzy OPF acts like a \emph{generalization} of the naive OPF classifier, i.e., Fuzzy OPF converges to the regular OPF when $\sigma=1$ in Equation~\ref{e.fuzzyMembership}. Besides, Table~\ref{t.results} strengths the hypothesis that OPF performs better than SVM over low-dimensional problems. In this case, OPF outperformed SVM in most of the experiments concerning the synthetic datasets.

\subsection{Computational Burden}
\label{ss.computational_birden_training}

Table~\ref{t.training} presents the computational burden (in seconds) demanded by each technique. Although Fuzzy OPF obtained a slower performance than the regular OPF, which is expected since the Fuzzy OPF requires performing a clustering step to calculate the membership before classification, overcoming such a constraint does not represent a massive effort. Such an issue can be easily surmounted using a parallelized implementation of the minimum graph cut employed for searching $k^{ast}$ (Section~\ref{ss.unsupervisedOPF}). Besides, Fuzzy OPF is fastest than SVM in $9$ out of $12$ datasets. 

\begin{table}[!htb]
\caption{\label{t.training}Average computational burden (in seconds) concerning the training step.}
\begin{center}
\renewcommand{\arraystretch}{1.5}
\setlength{\tabcolsep}{6pt}
\resizebox{\columnwidth}{!}{
\begin{tabular}{l|c|c|c|c}
\hhline{-|-|-|-|-|}
\hhline{-|-|-|-|-|}
\hhline{-|-|-|-|-|}
\cline{3-5}
\multicolumn{1}{c|}{\cellcolor[HTML]{EFEFEF}{\textbf{Dataset}}}  & \cellcolor[HTML]{EFEFEF}{\textbf{OPF Classifier}} & \cellcolor[HTML]{EFEFEF}{\textbf{Fuzzy-OPF}} & \cellcolor[HTML]{EFEFEF}{\textbf{SVM}} & \cellcolor[HTML]{EFEFEF}{\textbf{Bayes}}\\ \hline
\multicolumn{1}{c|}{Boat} & 0.00062 & 0.00583 & 0.47469 & 0.00295\\ \hline
\multicolumn{1}{c|}{Cone-Torus}  & 0.00501 & 0.20308 & 1.40612 & 0.00630\\ \hline
\multicolumn{1}{c|}{Four-Class}  & 0.02300 & 0.50150 & 4.80686 & 0.00887\\ \hline\hline
\multicolumn{1}{c|}{Data1}  & 0.05812 & 0.97826 & 0.62680 & 0.01304\\ \hline
\multicolumn{1}{c|}{Data2}   & 0.00267 & 0.07225 & 0.00660 & 0.00426\\ \hline
\multicolumn{1}{c|}{Data3}  & 0.00356 & 0.14377 & 0.00232 & 0.00497\\ \hline\hline
\multicolumn{1}{c|}{Thyroid } & 2.05512 & 12.37519 & 602.57032 & 0.17410\\ \hline
\multicolumn{1}{c|}{Breast Tissue} & 0.00037 & 0.01364 & 0.06333 & 0.00434\\\hline
\multicolumn{1}{c|}{Landsat Satellite}& 1.43291 & 9.90451 & 17.84326 & 0.20642 \\ \hline\hline
\multicolumn{1}{c|}{MPEG-7 BAS}  & 0.31309 & 4.79085 & 0.57394 & 0.28658\\ \hline
\multicolumn{1}{c|}{Elec. Ind. Prof.}   & 0.33162 & 3.09570 & 287.91745 & 0.04221\\ \hline
\multicolumn{1}{c|}{Elec. Com. Prof.} & 0.79331 & 5.78810 & 750.65645 & 0.06311\\ 
\hhline{-|-|-|-|-|}
\hhline{-|-|-|-|-|}
\hhline{-|-|-|-|-|}
\end{tabular}}
\end{center}
\end{table}

\subsection{Fine-Tuning Hyperparameters}
\label{ss.validation}

The main drawback of Fuzzy OPF when compared to the standard supervised OPF, at first glance, is related to a proper selection of its hyperparameters. However, since Fuzzy OPF is a generalization of naive OPF, setting up the $\sigma$ hyperparameter to one leads the technique to converge to the standard OPF, leaving $k_max$ practically irrelevant in such a case.

\begin{figure*}[!htb]
  \centerline{
    \begin{tabular}{ccc}
	\includegraphics[width=6cm]{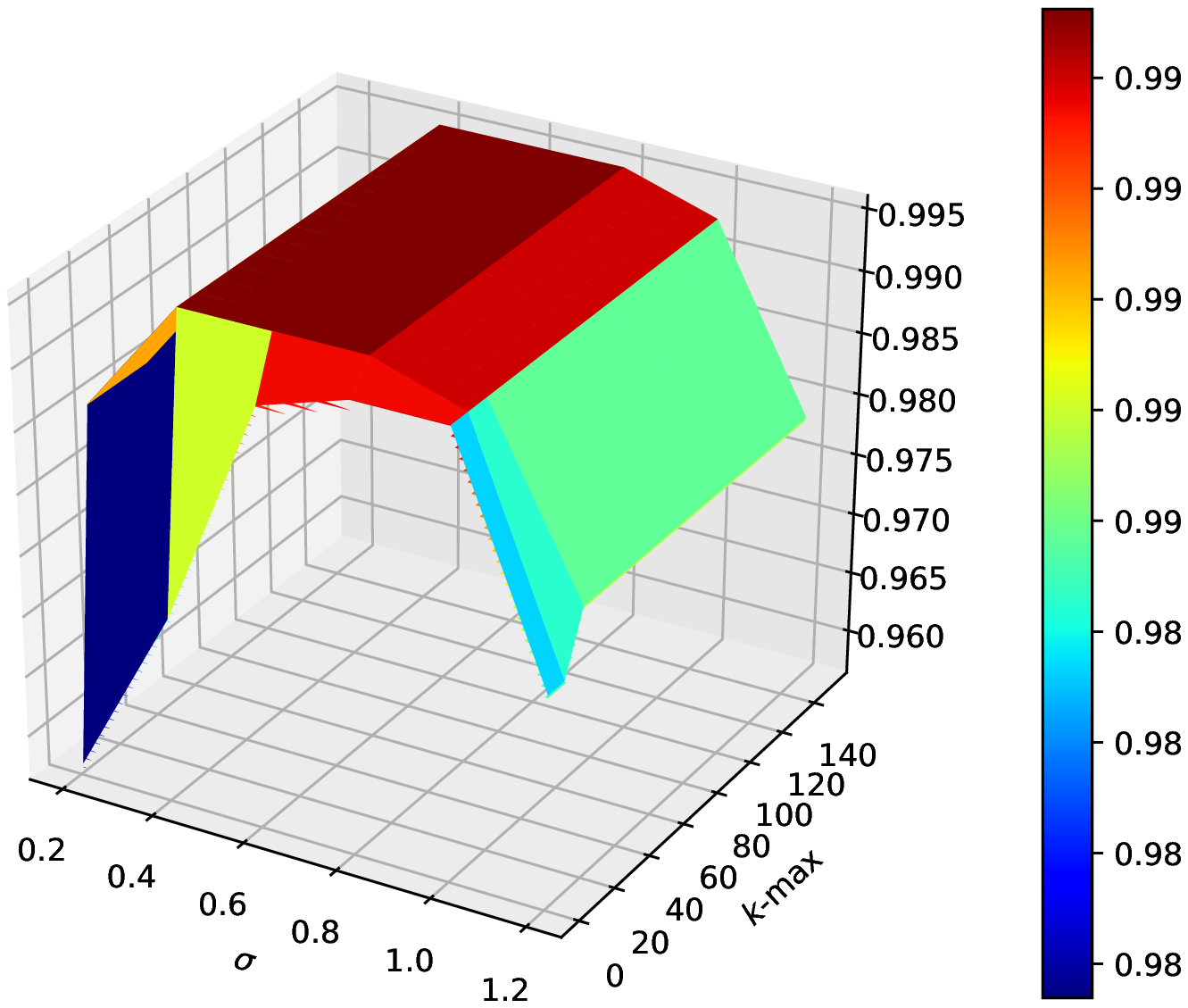} &
	\includegraphics[width=6cm]{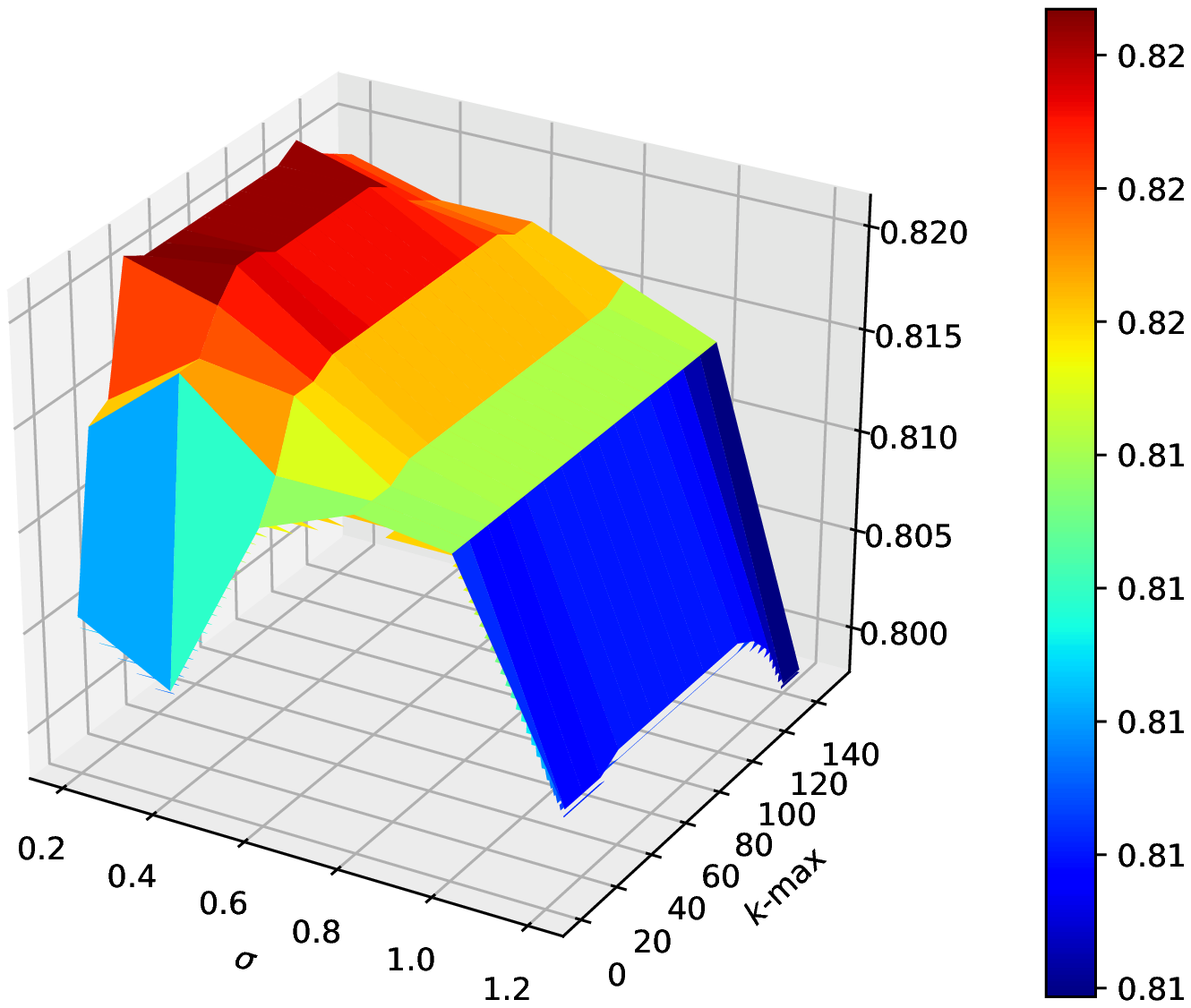}  &
	\includegraphics[width=6cm]{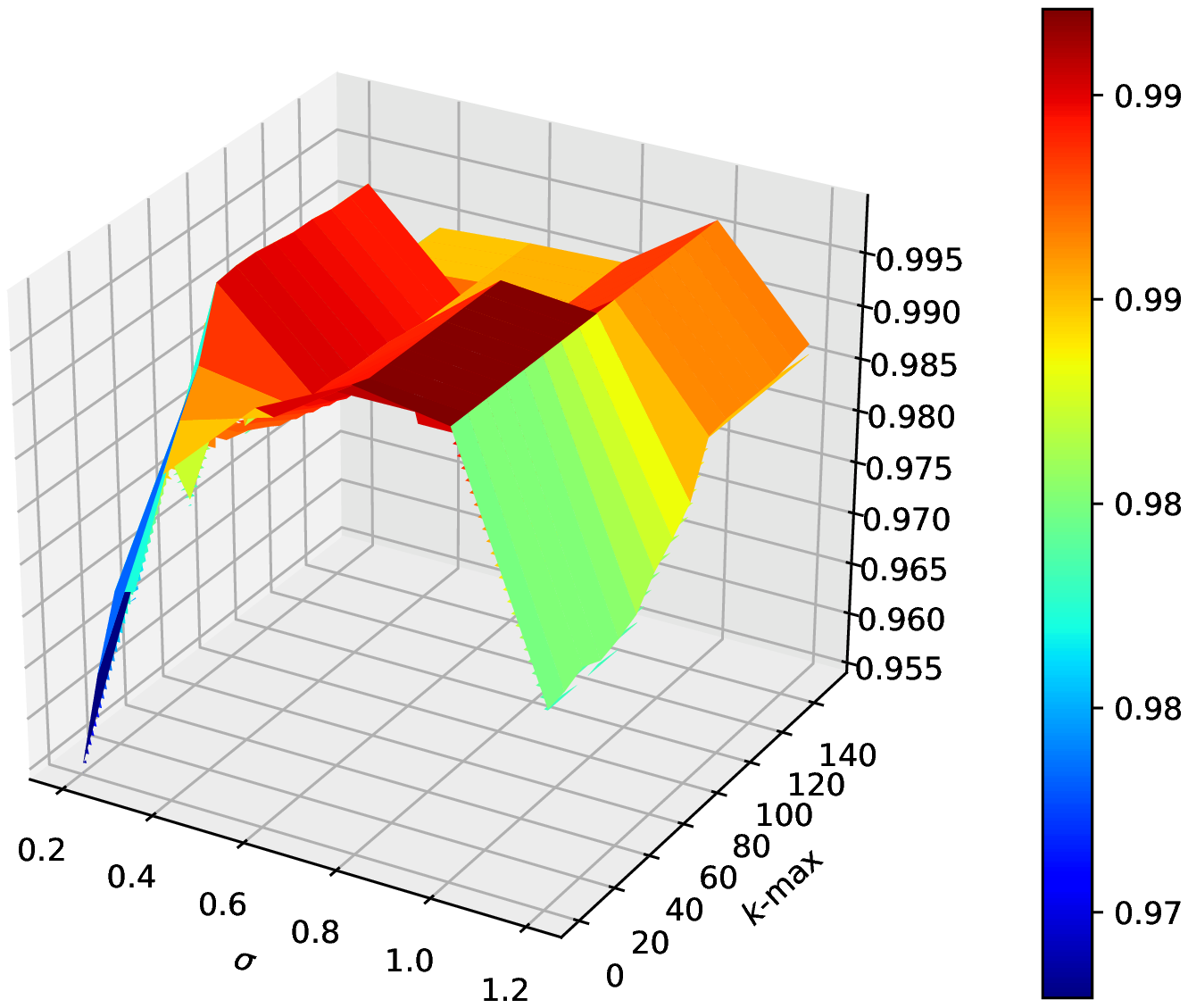} \\
	(a) & (b)  & (c)
    \end{tabular}}	
      \centerline{
    \begin{tabular}{ccc}
	\includegraphics[width=6cm]{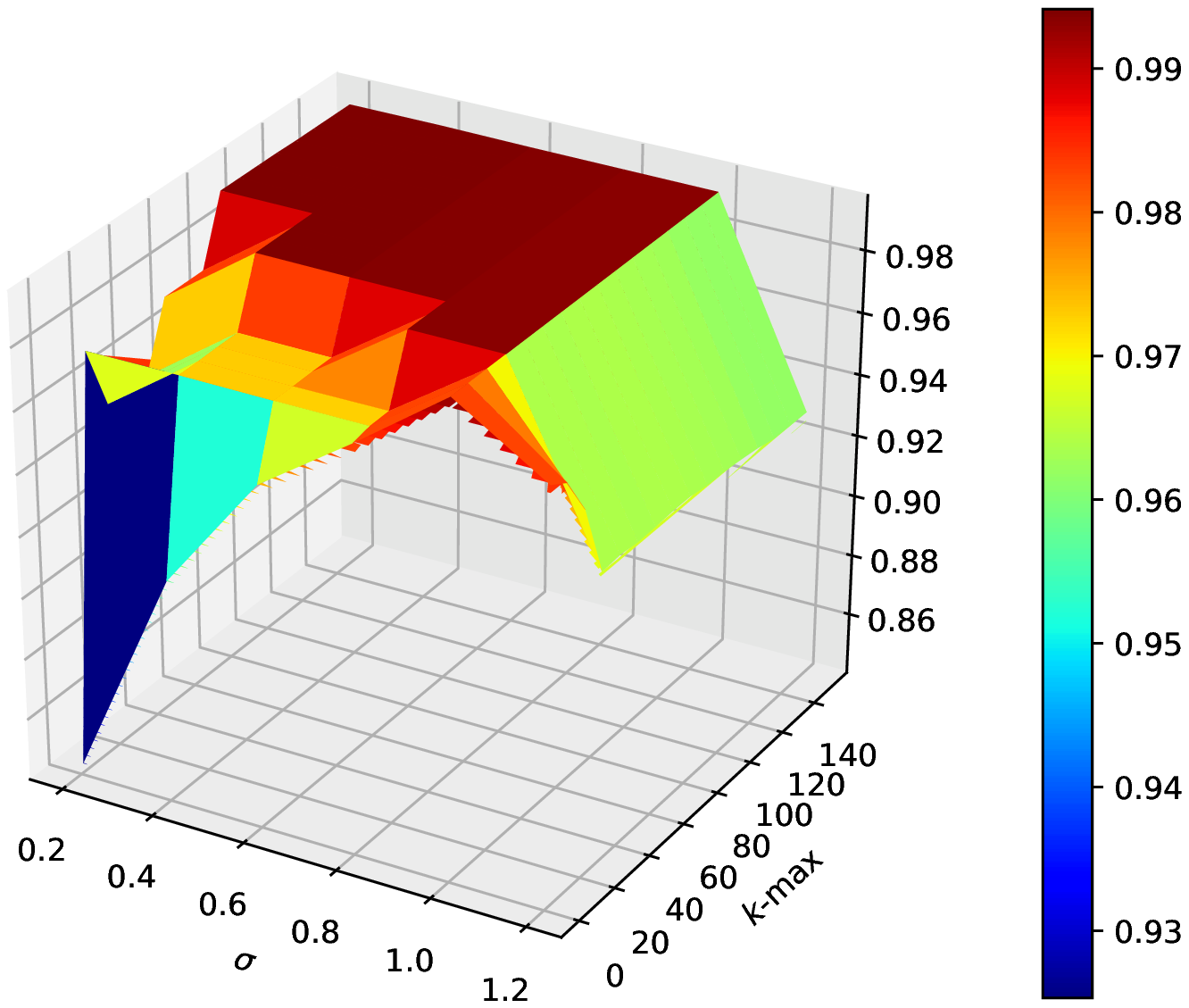} &
	\includegraphics[width=6cm]{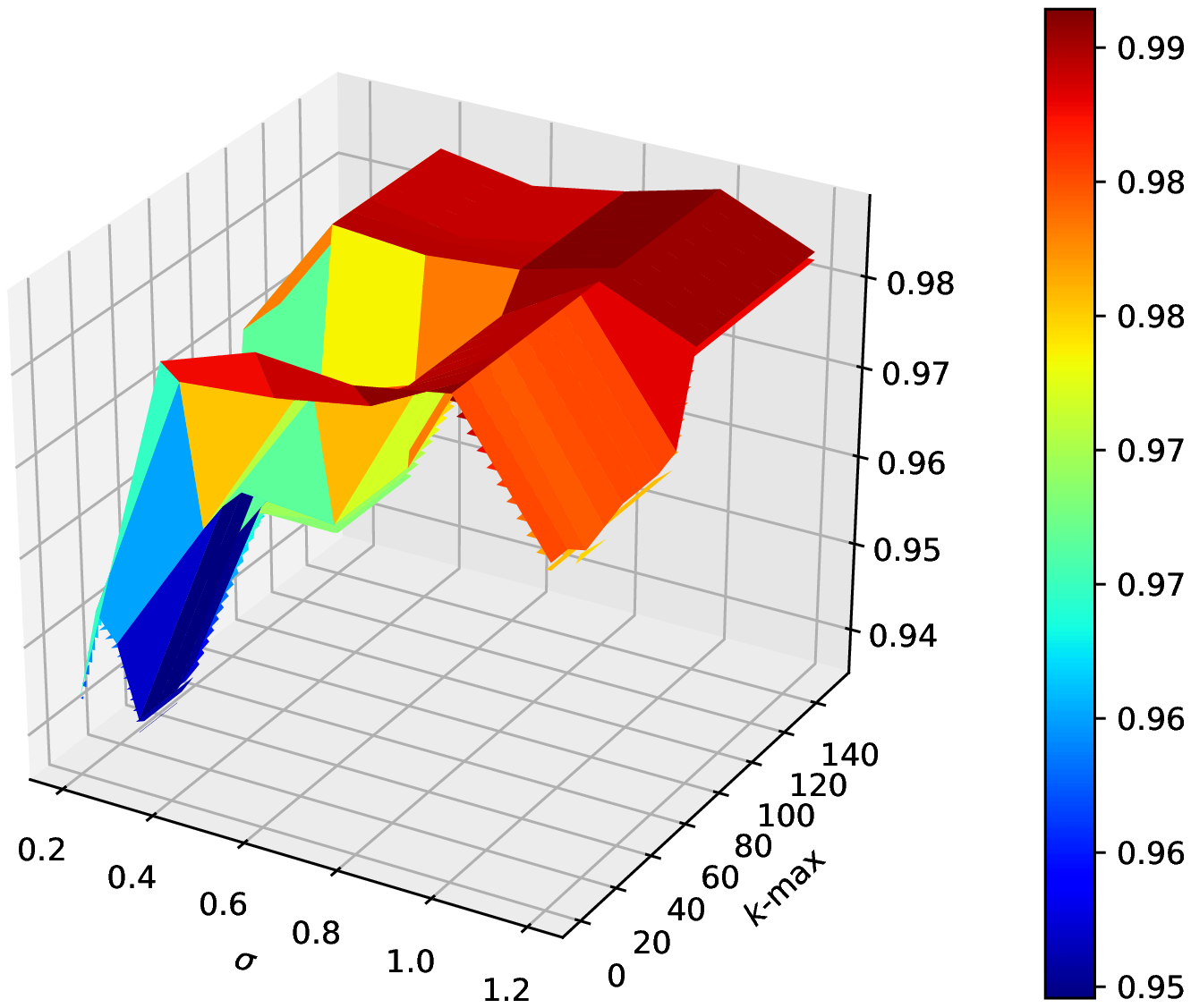} &
	\includegraphics[width=6cm]{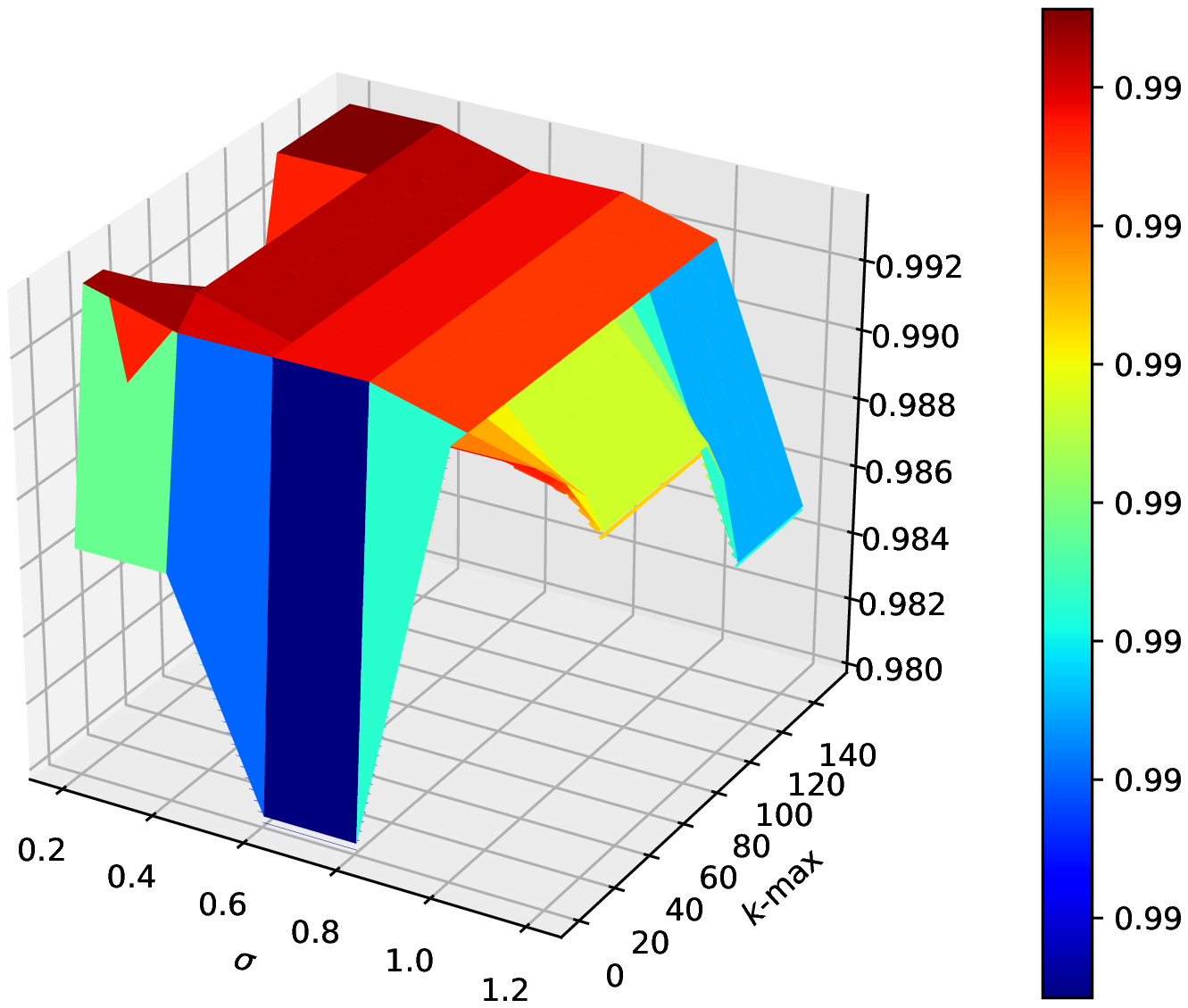}\\
	(d) & (e)  & (f) 
    \end{tabular}}
	\centerline{
    \begin{tabular}{ccc}
	\includegraphics[width=6cm]{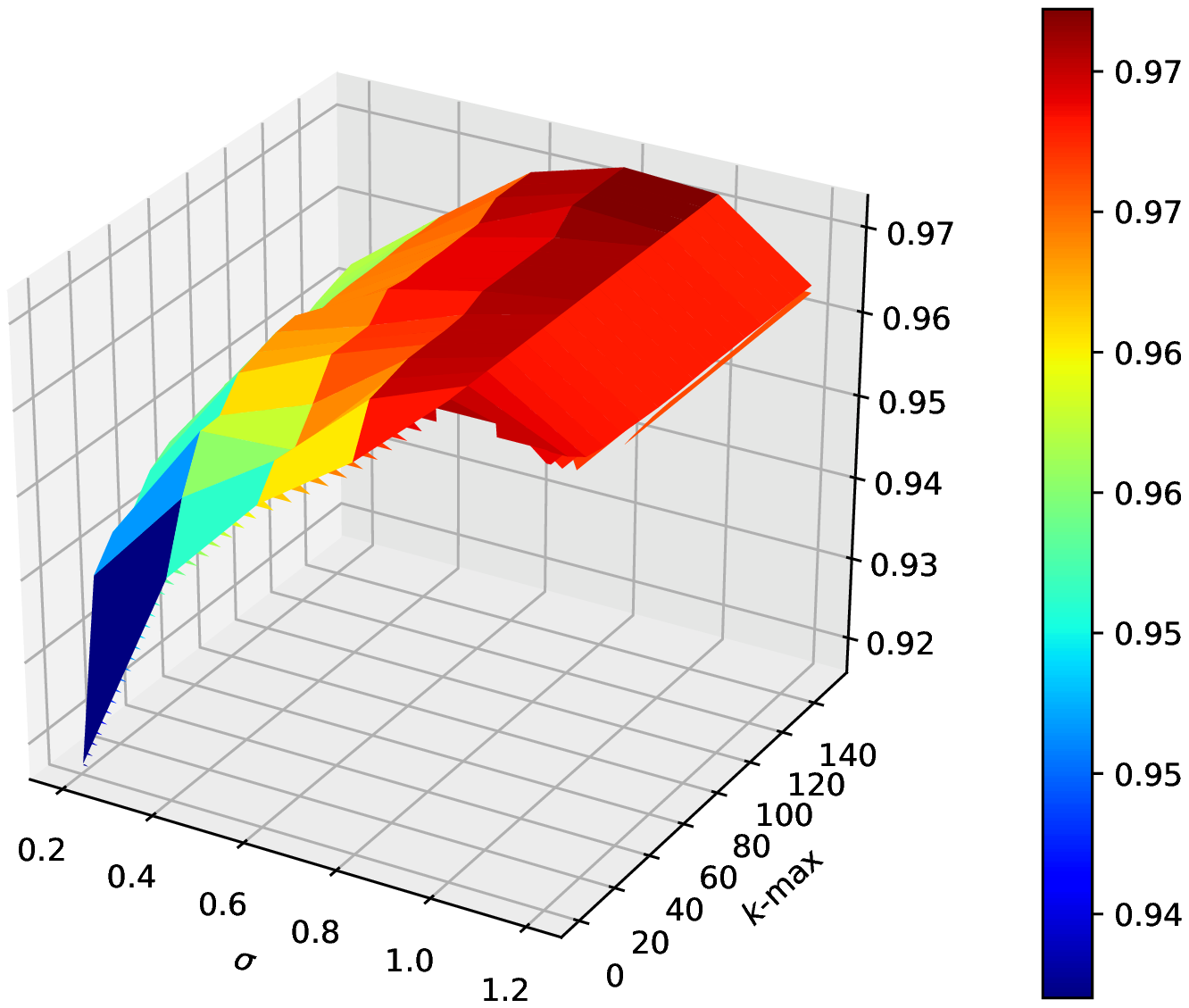} &
	\includegraphics[width=6cm]{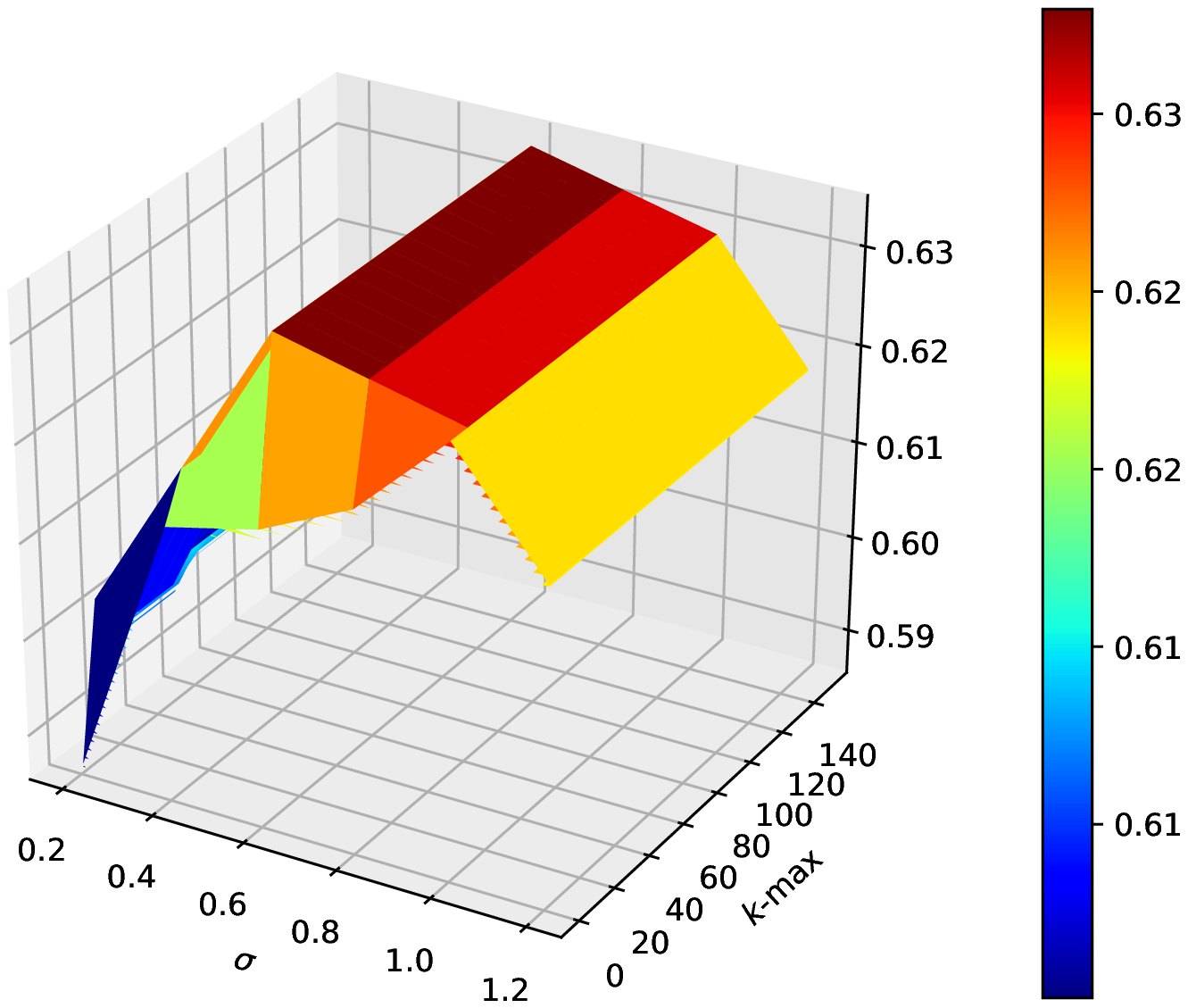} &
	\includegraphics[width=6cm]{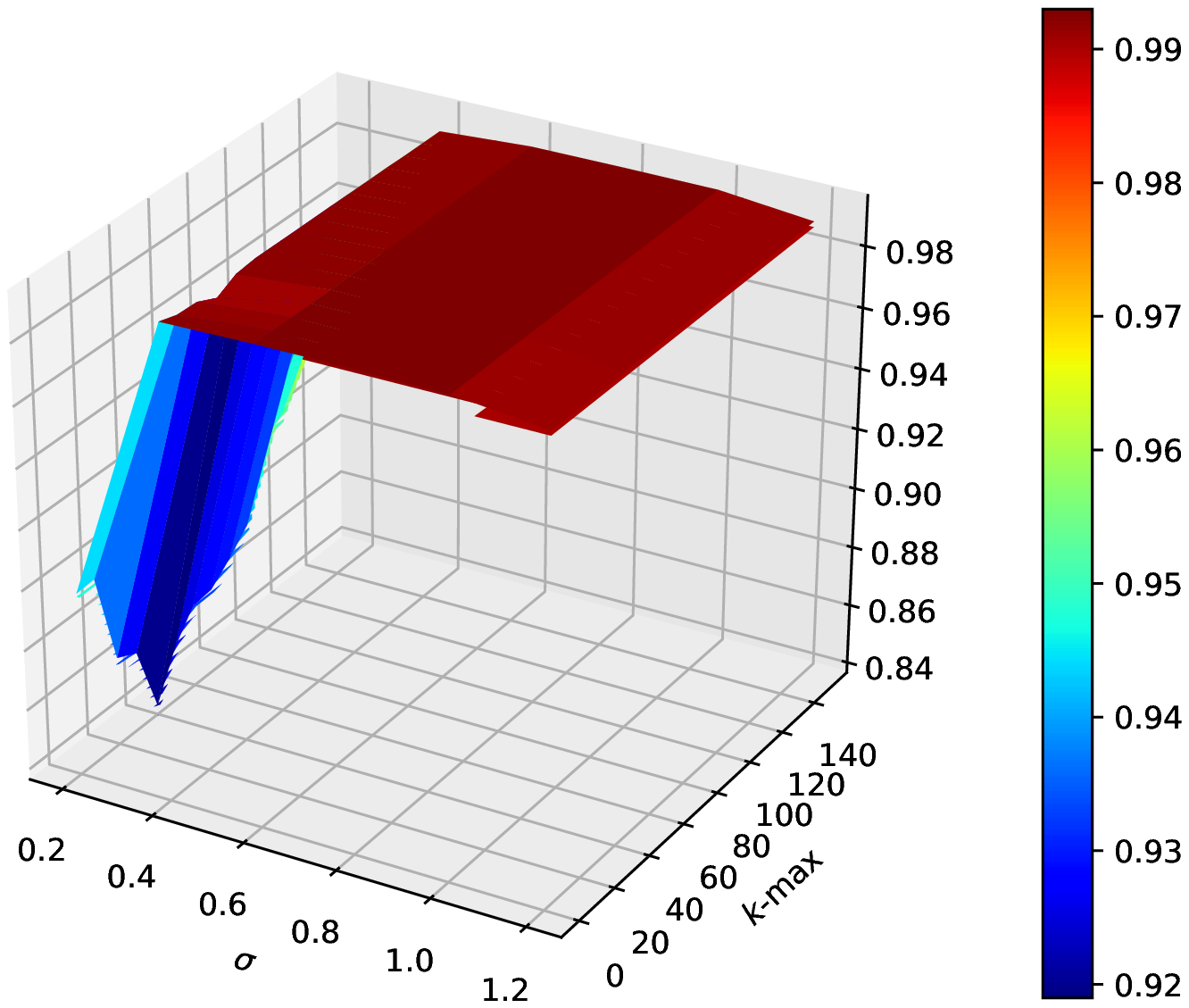}  \\ 
	(g) & (h)  & (i) 
    \end{tabular}}	
    	\centerline{
    \begin{tabular}{ccc}
	\includegraphics[width=6cm]{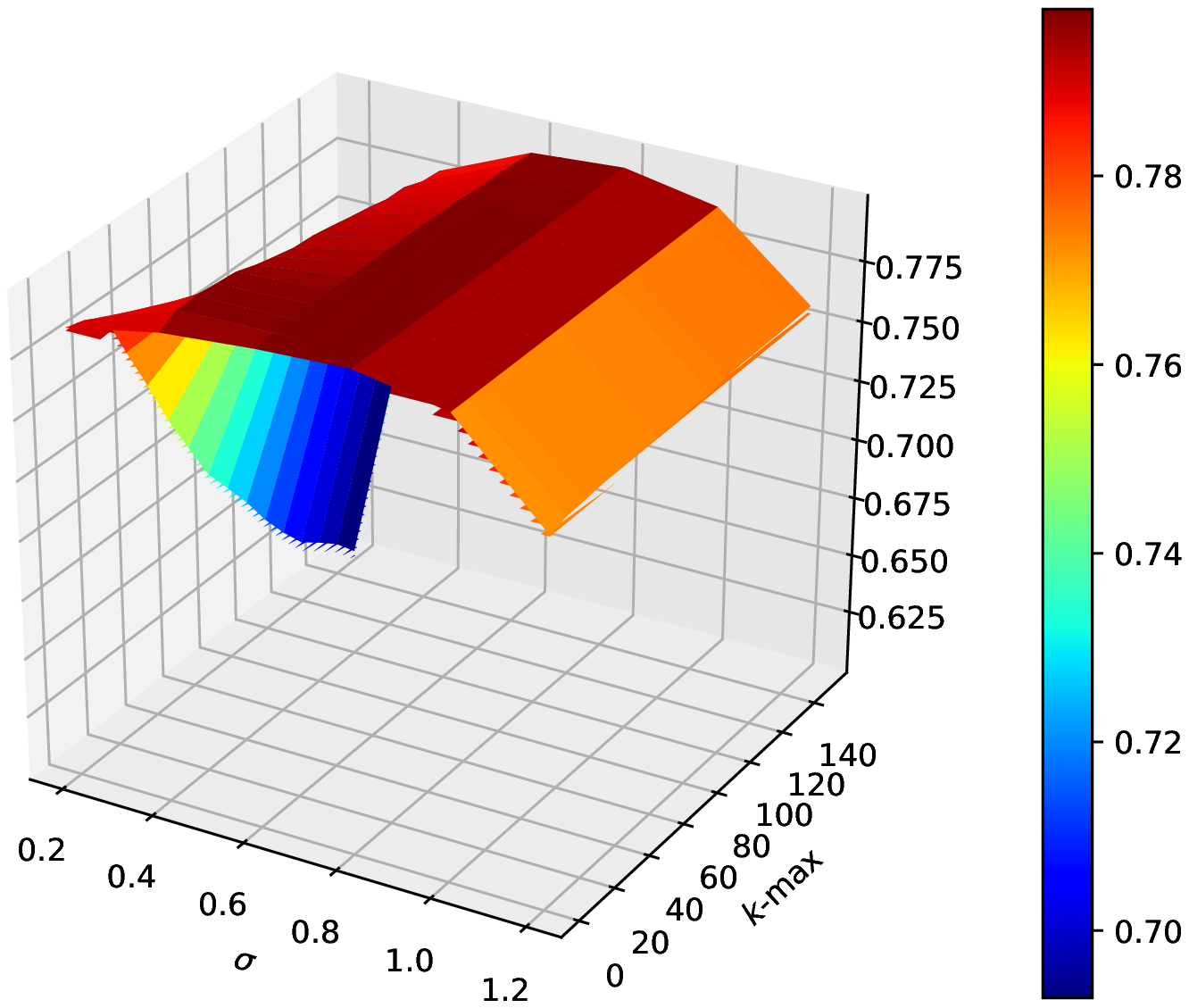} & 
	\includegraphics[width=6cm]{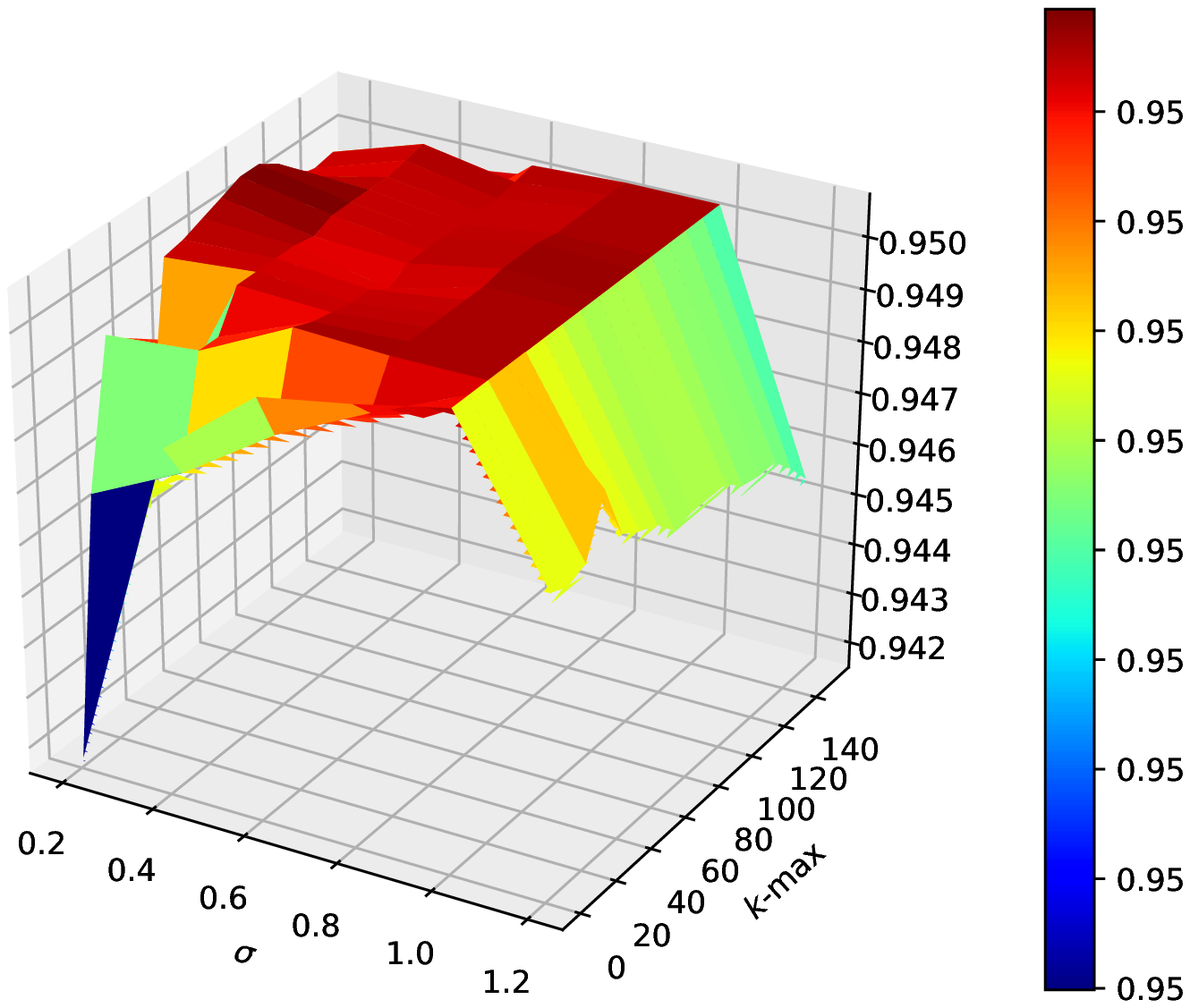}& 
	\includegraphics[width=6cm]{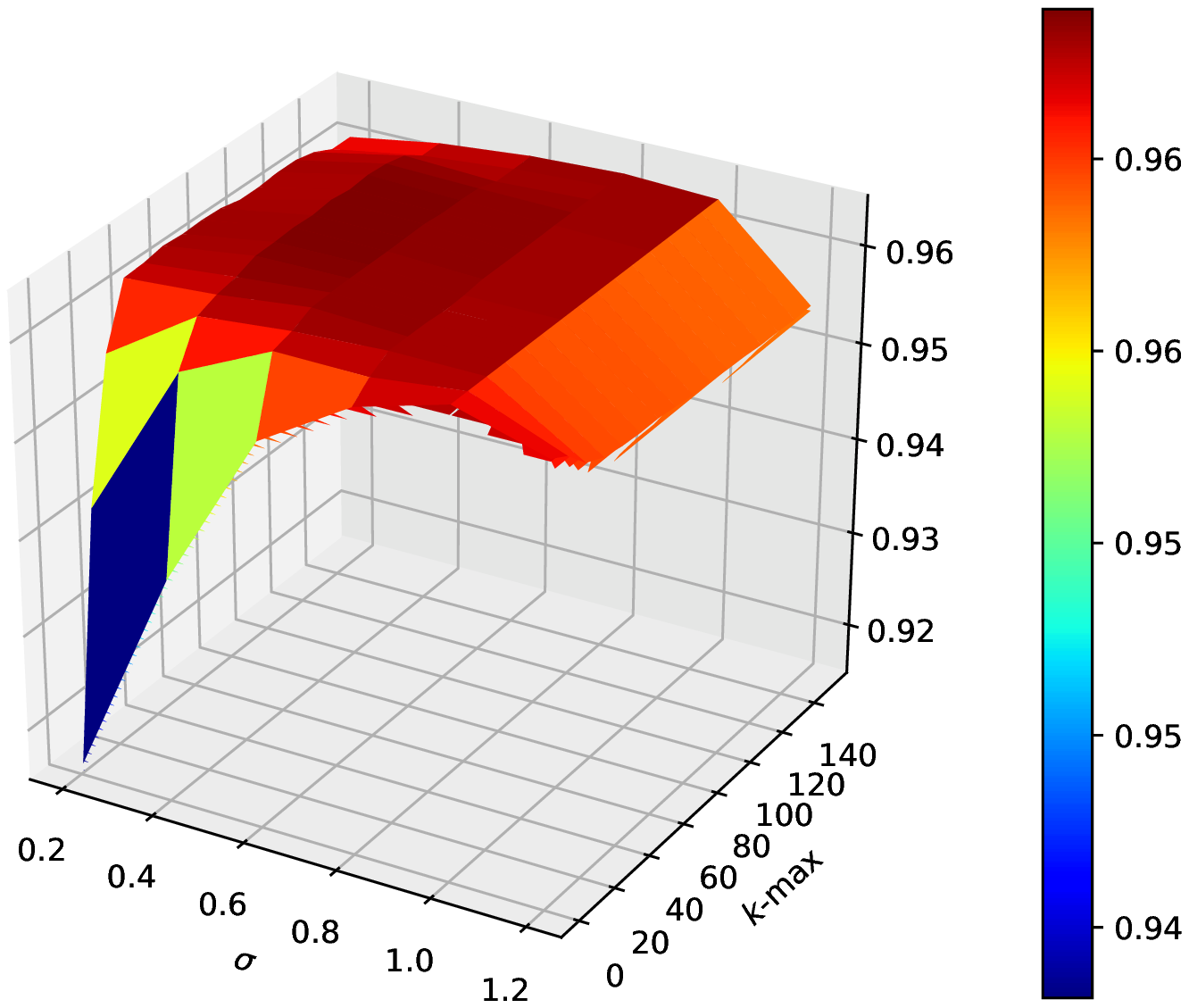} \\ 
	(j) & (k) & (l) 
    \end{tabular}}	
    \caption{Fuzzy OPF hyperparameter fine-tuning using a grid-search concerning: (a) Boat, (b) Cone-Torus, (c) Four-class, (d) Data1, (e) Data2, (f) Data3, (g) Thyroid, (h) Breast Tissue, (i) Landsat Satellite, (j) MPEG-$7$ BAS, (k) Electric Industrial Profiles, and (l) Electric Commercial Profiles datasets. Heater colors stand for higher accuracies.}
  \label{f.3dGraph}
\end{figure*}

Figure~\ref{f.3dGraph} depicts the fitness landscape for the hyperparameter fine-tuning procedure. One can observe that Fuzzy OPF does not require an accurate and sensible selection of its hyperparameters, i.e., a large portion of the area covers the best possible combinations of $\sigma$ and $k_{max}$ (red dark areas). In a nutshell, a random selection of such hyperparameters would probably provide satisfactory results in many cases.

\subsection{Discussion}
\label{ss.discussion}

One of the main problems related to the Optimum-Path Forest classifier concerns the so-called ``tie-zones", i.e., when two or more samples offer the very same optimum-path cost to a given sample. Standard OPF employs the FIFO (first-in-first-out) policy, i.e., the sample that offers the optimum-path cost first will be one that conquers others. 

However, we have observed that datasets that face too many overlapped areas among different classes pose a more significant challenge to any pattern classifier. Regarding the OPF technique, the conquering process in such areas is quite competitive, which means higher classification errors. In this context, it is pretty much usual a given sample be conquered by another one from a different class (i.e., a misclassification). 

In most cases, we have observed that the sample with the second best cost is the one labeled with the correct class. In this paper, we showed we could overcome such a shortcoming using memberships computed during the training step. Such fuzzy values are used to weight the optimum-path cost, which now considers such uncertainty. Notice that we are also aware that the path-cost function may not be \emph{smooth} anymore~\cite{FalcaoPAMI:04}, but in practice, such a circumstance does not interfere negatively in the process.

\begin{figure}[!h]
  \centerline{
   \begin{tabular}{c}   
	\includegraphics[width=6.3cm]{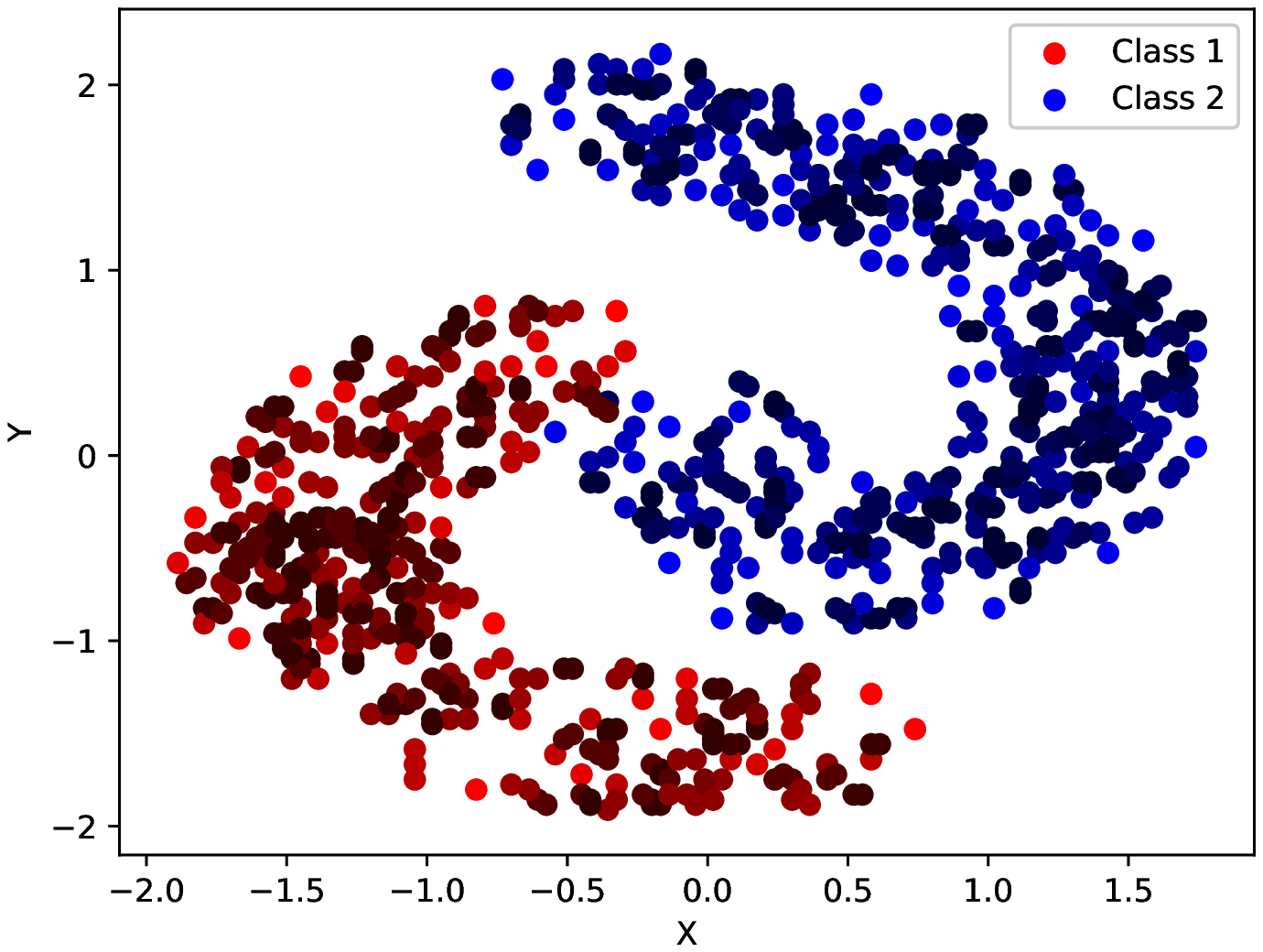} \\
	(a)
	 \end{tabular}}
	\centerline{
	\begin{tabular}{c}   
	\includegraphics[width=6.3cm]{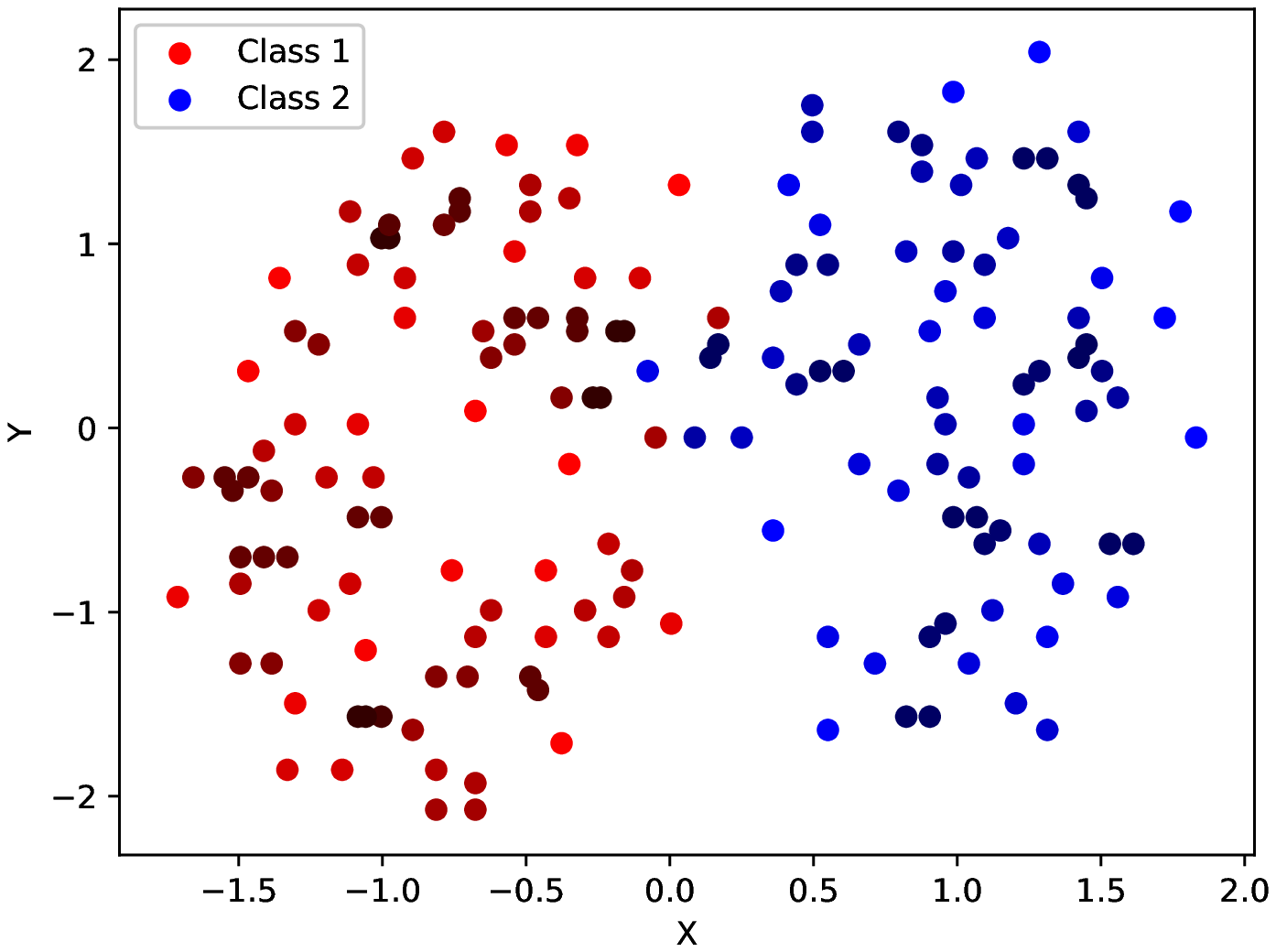} \\
	(b)
    \end{tabular}}	
    \caption{Membership values for all data points concerning: (a) Data1 and (b) Data2 datasets. Darker points stand for higher membership values.}
  \label{f.fuzzy_figures}
\end{figure}

We also conducted an extra round of experiments to show the rationale behind the proposed approach. Figures~\ref{f.fuzzy_figures}a and~\ref{f.fuzzy_figures}b display the membership values for all data points concerning Data1 and Data2 datasets, respectively. Notice that darker points stand for higher membership values. More representative samples (i.e., higher membership values) tend to be located at the center of the clusters, meanwhile points far away mind to be less important for training purposes. It is worth noting that unsupervised OPF finds clusters on-the-fly, i.e., there is no need to know the number of clusters beforehand. Such a skill is interesting since the number of clusters is usually much greater than the number of classes. This can explain why one can have several darker points at close/distinct locations.

\section{Conclusions}
\label{s.conclusions}

This paper proposed a variant of na\"ive supervised OPF that considers fuzzy information for classification purposes. Experiments conducted over ten public and two private datasets allowed us to draw some interesting conclusions: (i) Fuzzy OPF obtained the best results, according to the Wilcoxon signed-rank test, in eleven out of twelve datasets, achieving the highest results in eight of them; and (ii) Fuzzy OPF acts like a generalization and converges to the standard OPF when $\sigma = 1$.

Fuzzy OPF also provides the opportunity to outperform the naive OPF with a proper selection of the hyperparameters, although its demands a bit more computational resources since it requires a previous clustering step before classification. However, such an issue is easily overcome using parallel computing. Besides, the area covered by the combination of the best hyperparameters occupies a large portion of the search space mostly, which means a random selection of such values may provide satisfactory results. Finally, OPF-based techniques outperformed SVM in low-dimensional problems.

Regarding future works, we plan to employ meta-heuristic optimization techniques to fine-tune Fuzzy OPF hyperparameters. Additionally, we intend to propose a comparison among several unsupervised clustering techniques to calculate the membership values. 

\section*{Acknowledgements}
The authors are grateful to CNPq 304315/2017-6, 427968/2018-6, 430274/2018-1, and 307066/2017-7 grants, CAPES, FAPESP grants 2013/07375-0, 2014/12236-1, 2017/25908-6, 2018/21934-5, and 2016/19403-6, as well as the National Natural Science Foundation of China under Grant 61976120, the Natural Science Foundation of Jiangsu Province under Grant BK20191445, the Six Talent Peaks Project of Jiangsu Province under Grant XYDXXJS-048, and sponsored by Qing Lan Project of Jiangsu Province.

\ifCLASSOPTIONcaptionsoff
  \newpage
\fi

\bibliographystyle{IEEEtran}
\bibliography{./references}

\end{document}